\RequirePackage{fix-cm}
\documentclass{svjour3}                     % onecolumn (standard format)
\smartqed  % flush right qed marks, e.g. at end of proof
\usepackage{graphicx}
\usepackage{amssymb}
\usepackage{graphics}
\usepackage{graphicx}
\usepackage[intlimits]{amsmath}
\usepackage{amsfonts}
\usepackage{url}
\usepackage{theorem}
\usepackage{cite}
\usepackage{booktabs}
\usepackage{multirow}

\usepackage {longtable}
\usepackage{tikz}
\usetikzlibrary{patterns}
\usepackage{booktabs}
\usepackage{multirow}
\usepackage{caption}
\usepackage{subcaption}
\usepackage{times}
\usepackage[ruled,lined,boxed,vlined,commentsnumbered]{algorithm2e}
\SetKwFunction{FRecurs}{FnRecursive}%

\usepackage{nomencl}
\makeglossary

\newcommand\numberthis{\addtocounter{equation}{1}\tag{\theequation}}

\newcommand{\pfa}{\mathcal{P}}					% pareto front approximation
\usepackage{lscape}
\usepackage{adjustbox}
\usepackage{algorithmic}

\DeclareMathOperator\erf{erf}
\newcommand{\pdf}{\xi}

\usepackage{ifthen}
\usepackage{xcolor}
\usepackage{pgfplots}
\pgfplotsset{compat = 1.3}
\usepackage{tikz-3dplot}
\usepackage{pgfplots,tikz-3dplot}

\usetikzlibrary{calc,fadings,decorations.pathreplacing}
\tikzstyle{a}=[circle, inner sep=2pt,fill=red!50, draw]
\tikzstyle{b}=[circle, inner sep=2pt,fill=blue!50, draw]
\tikzstyle{c}=[circle, inner sep=2pt,fill=green!50, draw]
\tikzstyle{d}=[circle, inner sep=2pt,fill=yellow!50, draw]

\usepackage{mathtools}

\usepackage{lipsum}
\begin{document}

\title{Efficient Computation of Expected Hypervolume Improvement Using Box Decomposition Algorithms\footnote{This paper extends from a conference paper \cite{yang2017computing}.}
}

\titlerunning{Computation of Expected Hypervolume Improvement Using Box Decomposition}        % if too long for running head

\author{Kaifeng Yang${}^1$         	\and
        Michael Emmerich${}^1$     \and
        Andr{\'e} Deutz${}^1$      	\and
   %     Carlos M. Fonseca${}^2$    
        Thomas B{\"a}ck${}^1$      	
}


\institute{K. Yang, A. Deutz, M. Emmerich and T. B{\"a}ck \at
			  	LIACS, Leiden University,\\
              	Niels Bohrweg 1, \#164\\
              	2333 CA Leiden, The Netherlands\\
              	Tel.: +31-(0)71-5276435\\
              	Fax: +31-(0)71-5276985\\
              	\email{\{k.yang, a.h.deutz,  m.t.m.emmerich,t.h.w.baeck\}@liacs.leidenuniv.nl}           %  \\
%             	\emph{Present address:} of F. Author  %  if needed
%    \and
%           C. M. Fonseca\at
%           		CISUC, Department of Informatics Engineering, University of Coimbra\\
%           		Polo II, Pinhal de Marrocos, \\
%           		3030-290 Coimbra, Portugal \\
%           		\email{cmfonsec@dei.uc.pt}    
}

\date{Received: date / Accepted: date}

\maketitle

\begin{abstract}
In the field of multi-objective optimization algorithms, multi-objective Bayesian Global Optimization (MOBGO) is an important branch, in addition to evolutionary multi-objective optimization algorithms (EMOAs). 
MOBGO utilizes Gaussian Process models learned from previous objective function evaluations to decide the next evaluation site by maximizing or minimizing an infill criterion. A commonly used criterion in MOBGO is the Expected Hypervolume Improvement (EHVI), which shows a good performance on a wide range of problems, with respect to exploration and exploitation. 
However, so far, it has been a challenge to calculate exact EHVI values efficiently.
This paper proposes an efficient algorithm for the exact calculation of the EHVI for in a generic case. This efficient algorithm is based on partitioning the integration volume into a set of axis-parallel slices. Theoretically, the upper bound time complexities can be improved from previously $O (n^2)$ and $O(n^3)$, for two- and three-objective problems respectively, to $\Theta(n\log n)$, which is asymptotically optimal. This article generalizes the scheme in higher dimensional cases by utilizing a new hyperbox decomposition technique, which is proposed by D{\"a}chert et al., EJOR, 2017. It also utilizes a generalization of the multilayered integration scheme that scales linearly in the number of hyperboxes of the decomposition. 
The speed comparison shows that the proposed algorithm in this paper significantly reduces computation time. Finally, this decomposition technique is applied in the calculation of the Probability of Improvement (PoI).
\end{abstract}
%replaces exact objective function evaluations by
%using predictions from the so-called Kriging or Gaussian process models

\section{Introduction}\label{sec:intro}
In multi-objective design optimization, the objective function evaluations are generally computationally costly, mainly due to the long convergence times of simulation models. A simple and common remedy to this problem is to use a statistical model learned from previous evaluations as the fitness function, instead of the 'true' objective function. This method is also known as Bayesian Global Optimization (BGO) \cite{Mockus1978}. In BGO, a Gaussian Process (GP) model is used as a statistical model. In each iteration, the algorithm evaluates a new solution and updates the Gaussian Process model. A new solution is chosen by the score of an infill criterion, given a statistical model. For multi-objective problems, the family of these algorithms is called \emph{Multi-objective Bayesian global optimization} (MOBGO). 
Compared to evolutionary multi-objective optimization algorithms (EMOAs),
MOBGO requires only a small budget of function evaluations to achieve a similar result with respect to hypervolume indicator, and it has already been used in real-world applications to solve expensive evaluation problems \cite{kyang2015cec}. 
%CH -- remove comma after evaluations in "statistical model learned from previous evaluations," above
%CH -- remove "the" from "been used in the real world applications" above, "real world" is being used as an adjective, "What kind of appplications?" = "real world" applications, "green" applications, etc.
% MOBGO requires only a small budget of function evaluations (less than 300) \cite{kyang2015cec} and it has already been used in the real world optimization to solve expensive evaluation problems. 
According to the authors' knowledge, BGO was used for the first time in the context of airfoil optimization in \cite{laniewski2010development}, and then applied in the field of biogas plant controllers \cite{gaida2014dynamic}, detection in water quality management \cite{zaefferer2013case}, machine learning algorithm configuration \cite{koch2015efficientnoise}, and structural design optimization \cite{shimoyama2013updating}. 

In the context of Bayesian global optimization, an infill or pre-selection criterion is used to evaluate how promising a new point is. In single-objective optimization, the \emph{Expected Improvement} (EI) is widely used as the infill criterion, and it was first introduced by Mockus et al. \cite{Mockus1978} in 1978. The EI exploits both the Kriging prediction and the variance in order to give a quantitative measure of a solution's improvement. Later, the EI became more popular due to the work of Jones et al \cite{jones1998efficient}. In MOBGO, a commonly used criterion is \emph{Expected Hypervolume Improvement} (EHVI), which is a straightforward generalization of the EI and was proposed by Emmerich \cite{emmerich2005single} in 2005. Compared with other criteria, EHVI leads to an excellent convergence to -- and coverage of -- the true Pareto front \cite{couckuyt2014fast, kyang2016cec}. Nevertheless, the calculation of EHVI itself so far has been time-consuming \cite{shimoyama2013kriging, zaefferer2013case, wagner2010expected,koch2015efficientnoise}, even in the 2-D case\footnote{In this paper, 2-D and 3-D represent two and three objective functions, respectively.}. Moreover, EHVI has to be computed many times by an optimizer to search for a promising solution in every iteration. For these reasons, a fast algorithm for computing EHVI is needed. 

The first method suggested for EHVI calculation was Monte Carlo integration and was proposed by Emmerich \cite{emmerich2005single, emmerich2006single}. This method is simple and straightforward. However, the accuracy of EHVI highly depends on the number of iterations. The first exact EHVI calculation algorithm in the 2-D case was derived in \cite{emmerich2011hypervolume}, with time complexity of $O(n^3\log n)$. Here, $n$ is the number of non-dominated points in the archive. The EHVI calculation algorithm in \cite{emmerich2011hypervolume} partitions an objective space into boxes and then calculates the EHVI by summing all the EHVI values of each box. 
Couckuyt et al. \cite{couckuyt2014fast} introduced an exact EHVI calculation algorithm (CDD13) for $d>2$ by representing a non-dominated space with three types of boxes, where $d$ represents the number of objective functions. 
The method in \cite{couckuyt2014fast} was also practically much faster than those discussed in \cite{emmerich2011hypervolume}, though a detailed complexity analysis was missing. Hupkens et al. \cite{hupkens2015faster} reduced the time complexity to $O(n^2)$ and $O(n^3)$ in the 2-D and 3-D cases, respectively. The algorithms in \cite{hupkens2015faster} improve the algorithms in \cite{emmerich2011hypervolume} by two ways: 1) only summing the EHVI values of each box in a non-dominated space; 2) reusing some intermediate integrations during the EHVI calculation.
The algorithms in \cite{hupkens2015faster} further improve the practical efficiency of EHVI on test data in comparison to \cite{couckuyt2014fast}. Recently, Emmerich et al. \cite{Michael2016book} proposed an asymptotically optimal algorithm with time complexity of $\Theta(n\log n)$ in the bi-objective case. More recently, Yang et al. proposed an asymptotically optimal algorithm with time complexity $\Theta(n\log n)$ in the 3-D case \cite{yang2017computing}. 
The algorithm, KMAC\footnote{KMAC stands for the authors' given names of the EHVI exact calculation algorithm.} in \cite{yang2017computing}, partitions a non-dominated space by slices linearly and re-derives the EHVI calculation formulas. 
%It has recently been shown in Emmerich et al.  \cite{Michael2016book} and, respectively, Yang et al. \cite{yang2017computing} that the EHVI can be computed by means of algorithms with asymptotically optimal time complexity in $\Theta(n\log n)$. 
%In these two papers, a new integration technique, which scales the number of divided non-dominated boxes linearly, was introduced in 2-D and 3-D cases. 
However, a generalization of this technique to more than three dimensions/objectives and the empirical testing of MOBGO algorithms on benchmark optimization problems, are still missing so far.

This paper mainly contributes to extending the state-of-the-art EHVI calculation methods into higher dimensional cases. The paper is structured as follows: Section \ref{sec:mobgo} introduces the nomenclature, Kriging, and the framework of MOBGO; Section \ref{sec:NotitionDefinition} provides some fundamental definitions used in this paper; Section \ref{sec:MCC} describes how to partition an integration space into (hyper)boxes efficiently, and how to calculate EHVI based on this partitioning method; Section \ref{sec:exp} shows experimental results of speed comparison and MOBGO based algorithms' performance on 10 well-known scientific benchmarks in 6- and 18-dimensional search spaces; Section \ref{sec:Conclusions} draws the main conclusions and discusses some potential topics for further research.

\section{Multi-objective Bayesian Global Optimization}\label{sec:mobgo}
A multi-objective optimization (MOO) problem is an optimization problem that involves multiple objective functions. A MOO problem can be formulated as:
\begin{align*}
	& ``\max" \big({y}_1(\mathbf{x}),{y}_2(\mathbf{x}),\cdots,{y}_d(\mathbf{x}) \big) \qquad  \qquad  \mathbf{x} \in \mathbb{R}^m \numberthis\\
%	& \text{subject to} \qquad 	
\end{align*}
where $d$ is the number of objective functions, ${y}_i(i=1,\cdots,d)$ are the objective functions, and a decision vector $\mathbf{x}$ is in an $m$-dimensional space.

% The basic idea of BGO is to use surrogate models based on Kriging or Gaussian processes. A surrogate model reflects the relationship between decision vectors and their corresponding 'true' objective values. The data of previous function evaluations will be used to build a surrogate model. For multi-objective problems, the family of these algorithms is called \emph{Multi-Objective Bayesian Global Optimization} (MOBGO).
\subsection{Notations}
The following table summarizes the notations used in this paper.
\begin{longtable}[h!]{p{50pt} p{50pt} p{200pt}}
\caption{Notations}\\
\hline
Symbol	& Type & Description \\
\hline
$m$											&  $\mathbb{N}^+$ 		& Dimension of a search space\\
$d$											&  $\mathbb{N}^+$ 		& Dimension of an objective space\\
$\boldsymbol\mu$							&  $\mathbb{R}^d$ 		& Mean values of predictive distribution\\
$\boldsymbol\sigma$							&  $(\mathbb{R}_0^+)^d$ & Standard deviations of predictive distribution\\
% $t$& $\mathbb{R}$ & \\
$\pfa$								        & $(\mathbb{R}^d)^n$ 	& A Pareto-front approximation set\\ %(Pareto front approximation in $t-1$)\\
$n$, $|\pfa|$								& $\mathbb{N}^+$ 		& Number of the non-dominated points in $\pfa$ \\
$\mathbf{y}^{(1)}, \dots, \mathbf{y}^{(n)}$ & $\mathbb{R}^d$ 		& The vectors in $\pfa$, where $\pfa = (\mathbf{y^{(1)},\dots,\mathbf{y^{(n)}}})$\\ 
$\mathbf{r}$								& $\mathbb{R}^d$ 		& Reference point\\
%$\mathrm{n_b}$& $\mathbb{N}$ & Number of axis-aligned rectangular boxes\\
%$\mathrm{B_1}, \dots, \mathrm{B_{n_b}}$& $(\mathbb{R}^d)^2$ & Rectangle hyper-boxes \\
$S_d$ 										& $(\mathbb{R}^d)^2$ 	& Integration slices in a $d$-dimensional space\\
$N_i$   				& $\mathbb{N}^+$ 		& Number of integration slice in $i$-dimensional case\\
$\mathbf{l}_d^{(1)}, \dots, \mathbf{l}_d^{(N_d)}$		& $\mathbb{R}^d$ 		& Lower bounds of integration slices \\
$\mathbf{u}_d^{(1)}, \dots, \mathbf{u}_d^{(N_d)}$		& $\mathbb{R}^d$ 		& Upper bounds of integration slices \\
$\mathbf{x}^t$								& $\mathbb{R}^d$ 		& A target solution in a search space\\
$\mathbf{x}^*$								& $\mathbb{R}^d$ 		& A promising solution in a search space\\	
$M_i$ 		& $i = 1, \cdots, d$ 	& The Kriging model for the $i$-th objective function\\
$g$ 			& $\mathbb{N}^+$			& Counter/Number of function evaluations \\	
$T_c$		& $\mathbb{N}^+$			& Termination criterion \\	
$\eta$		& $\mathbb{N}^+$			& The number of initialized solutions  \\
$D$ 			& $(\mathbb{R}^{(d+m)})^{g}$					& Training dataset for the Kriging models $M$\\
$\lambda_i$ & $i = 1, \cdots, d$ 	& The Lebesgue measure on $\mathbb{R}^i$ \\
%$\lambda_1$ & / 						& The Lebesgue measure on $\mathbb{R}$ \\
$\boldsymbol{\pdf}$ 		&   			& The multivariate independent normal distribution \\
$S_i$ 		& $i = 1, \cdots, d$ 	& The integration slices in an i-dimensional case \\ 
%$i$ 			& $i = 1, \cdots, d$ 	& \\
$y$ 										&  					& Objective functions \\
\hline
\label{tab:notation}
\end{longtable}
%In multi-objective optimization, the objective function evaluations are usually costly, and evolutionary multi-objective optimization algorithms are generally not efficient to solve these expensive function evaluation problems. This is because EMOAs typically require a large number of function evaluations. Furthermore, the Pareto optimality of the solutions cannot be guaranteed with fewer function evaluations. Therefore, EMOA is not recommended when function evaluations are expensive. In such cases, a better idea is perhaps to utilize information from all previous evaluations. This kind of algorithm is called \emph{Bayesian Global Optimization} (BGO), which was proposed by Mockus et al. in \cite{Mockus1978}.
\subsection{Kriging}
As a statistical interpolation method, Kriging is a Gaussian process based multivariate regression method. 
%Its evaluation cost is typically low compared to simulator based evaluations in design optimization \cite{li2008metamodel}. Kriging is a popular surrogate model to approximate noise-free data in computer experiments. 
Compared with simulator-based evaluations in design optimization, one prediction/evaluation of the Kriging model is typically cheap \cite{li2008metamodel}. Therefore, Kriging is widely used as a popular surrogate model to approximate noise-free data in computer experiments. 
Kriging models are fitted from previously evaluated points. 
%CH -- remove "the" "from Kriging models are fitted from the previously evaluated points." You are describing Kriging in general, so we are not talking about specific known points. "The" restricts your usage to known points in an actual instance of Kriging.
%Then, the Kriging models replace the real time-consuming simulation model \cite{sacks1989design} and act as the objective functions. 
Given a set of $n$ decision vectors $\mathbf{X}=\{\mathbf{x}^{(1)}, \mathbf{x}^{(2)}, \cdots , \mathbf{x}^{(n)}\},\mathbf{x}^{(i)} \in \mathbb{R}^m|_{i,=1,\cdots,n}$ in an $m$-dimensional search space, and associated function values $\mathbf{Y}(\mathbf{X}) =\big({y}(\mathbf{x}^{(1)}), {y}(\mathbf{x}^{(2)}), \cdots, {y}(\mathbf{x}^{(n)})\big)^{\top}$, Kriging assumes $\mathbf{Y}(\mathbf{X})$ to be a realization of a random process $Y$ of the following form \cite{chugh2017handling,jones1998efficient}:
\begin{align*}
Y(\mathbf{x}) = \mu(\mathbf{x}) + \epsilon(\mathbf{x}) \numberthis
\end{align*} 
where $\mu(\mathbf{x})$ is the estimated mean value over all given sampled points, and $\epsilon(\mathbf{x})$ is a realization of a Gaussian process with zero mean and variance $\sigma^2$. The regression part $\mu(\mathbf{x})$ approximates the function $Y$ globally and the Gaussian process $\epsilon(\mathbf{x})$ takes local variations into account. Opposed to other regression methods (such as support vector machine), Kriging/GP also provides an uncertainty qualification of a prediction. The correlation between the deviations at two decision vectors ($\mathbf{x}$ and $\mathbf{x'}$) is defined as:
\begin{align*}
Corr[\epsilon(\mathbf{x}),\epsilon(\mathbf{x'})] 
= R(\mathbf{x},\mathbf{x'})
= \prod_{i=1}^m R_i(x_i,x_i') \numberthis
\end{align*}
Here $R(.,.)$ is the correlation function, which decreases with the distance between two points. It is common practice to use a Gaussian correlation function (also known as a squared exponential kernel):
\begin{align*}
R(\mathbf{x},\mathbf{x'}) = \prod^m_{i=1} \text{exp}(-\theta_i(x_i - x'_i)^2) \qquad (\theta_i \geq 0) \numberthis
\end{align*}
where $\theta_i$ are parameters of the correlation model. They can be interpreted as a measurement of the variables' importance. 
The optimal $\boldsymbol{\theta} = (\theta_1^{opt}, \cdots, \theta_m^{opt})$ in the Kriging models are usually optimized by a continuous optimization algorithm. In this paper, the optimal $\boldsymbol{\theta}$ is optimized by the simplex search method of Lagarias et al. ($fminsearch$) \cite{lagarias1998fminsearch}, with the parameter of max function evaluations equal to 1000.
The covariance matrix can then be expressed through the correlation function:
\begin{align*}
Cov({\epsilon(\mathbf{x})}) = \sigma^2 \mathbf{\Sigma}, \qquad \text{where} \qquad \mathbf{\Sigma}_{i,j}=R(\mathbf{x_i},\mathbf{x_j}) \numberthis
\end{align*}
%$$Cov(\boldsymbol{\epsilon}) = \sigma^2 \mathbf{\Sigma}, \qquad \text{where} \qquad \mathbf{\Sigma}_{i,j}=R(\mathbf{x_i},\mathbf{x_j})$$

When $\mu(\mathbf{x})$ is assumed to be an unknown constant, the unbiased prediction is called ordinary Kriging (OK). In OK, the Kriging model determines the hyperparameters $\mathbf{\theta} = [\theta_1, \theta_2, \cdots, \theta_n]$ by maximizing the likelihood over the observed dataset. The expression of the likelihood function is:
\begin{align*}
L = -\frac{n}{2}\ln({\sigma}^2) - \frac{1}{2}\ln(|\mathbf{\Sigma}|) \numberthis
\end{align*}
The maximum likelihood estimates of the mean $\hat{\mu}$ and the variance $\hat{\sigma}^2$ are given by:
\begin{align*}
&\hat{\mu} = \frac{\mathbf{1}^{\top}_n \mathbf{\Sigma}^{-1} \mathbf{y}}{\mathbf{1}^{\top}_n \mathbf{\Sigma}^{-1} \mathbf{1}_n} \numberthis\\
&\hat{\sigma}^2 = \frac{1}{n}(\mathbf{y} - \mathbf{1}_n\hat{\mu})^{T}\mathbf{\Sigma}^{-1}(\mathbf{y}-\mathbf{1}_n\hat{\mu}) \numberthis
\end{align*}

Then the predictor of the mean and the variance at a target point $\mathbf{x}^t$ can be derived. They are shown in \cite{jones1998efficient}:
\begin{align*}
&\mu(\mathbf{x}^t) = \hat{\mu} + \mathbf{c}^{\top} \mathbf{\Sigma}^{-1} (\mathbf{y} - \hat{\mu}\mathbf{1}_n) \numberthis \\
&\sigma^2(\mathbf{x}^t) = \hat{\sigma}^2[1 - \mathbf{c}^{\top} \mathbf{\Sigma}^{-1}\mathbf{c} + \frac{1-\mathbf{c}^{T}\Sigma^{\top}\mathbf{c}}{\bf{1}_n^{\top}\Sigma^{-1}\mathbf{1}_n}] \numberthis
\end{align*}
where $\mathbf{c} = (Corr[y(x^t),y(x_1)], \cdots, Corr[y(x^t),y(x_n)])^{\top}$.
\subsection{Structure of MOBGO}
In MOBGO, it is assumed that $d$ objective functions are mutually independent in an objective space. Each objective function is approximated by a Kriging model individually, based on the $\eta$ existing evaluated data $D =  \big( (\mathbf{x}^{(1)},  \mathbf{y}^{(1)}=y(\mathbf{x}^{(1)})), \dots, (\mathbf{x}^{(\eta)}, \mathbf{y}^{(\eta)}=y(\mathbf{x}^{(\eta)})) \big )$.
Each Kriging model is a one-dimensional normal distribution, with a mean $\mu$ and a standard deviation $\sigma$. 
Given a target solution $\mathbf{x}^{t}$, the Kriging models can predict the multivariate outputs by means of an independent joint normal distribution with means  $\mu_1$, $\dots$, $\mu_d$ and standard deviations $\sigma_1$, $\dots$, $\sigma_d$. 
These predictive means and standard deviations are used to calculate the score of an infill criterion, which can quantitatively measure how promising the target point $\mathbf{x}^{t}$ is when compared with the current Pareto-front approximation set.  
A promising solution $\mathbf{x}^*$ can be found by maximizing/minimizing\footnote{It depends on which criterion is chosen.} the score of the infill criterion. 
%To calculate the score of the infill criterion is predicted by the Kriging models, instead of calculated by the 'true' objective functions. 
Then, this promising solution $\mathbf{x}^*$ is evaluated by the 'true' objective functions, and both the dataset $D$ and the Pareto-front approximation set $\pfa$ are updated. 

The basic structure of the MOBGO algorithm is shown in Algorithm \ref{alg:mobgo}. It mainly contains three parts: initialization of a sampling dataset, searching for an optimal solution and updating the Kriging models, and returning the Pareto-front approximation set $\pfa$. 

\begin{algorithm}[!h]
\caption{\textbf{MOBGO algorithm}}
\begin{algorithmic}[1]
\REQUIRE {Objective functions $y$, initialization size $\eta$, termination criterion $T_c$}
\ENSURE  {Pareto-front approximation $\pfa$}
\STATE Initialize $\eta$ solutions $\{\mathbf{x}^{(1)}, \cdots, \mathbf{x}^{(\eta)}\}$ using LHS algorithm; 
\STATE Evaluate the initial set of $\eta$ points: $( \mathbf{y}^{(1)}=y(\mathbf{x}^{(1)}),\dots, \mathbf{y}^{(\eta)}=y(\mathbf{x}^{(\eta)}) )$;
%\STATE Store $(\mathbf{x}^{(1)}, \cdots, \mathbf{x}^{(\eta)})$ and $( \mathbf{y}^{(1)}=\mathbf{y}(\mathbf{x}^{(1)}),\dots, \mathbf{y}^{(\eta)}=\mathbf{y}(\mathbf{x}^{(\mu)}) )$ in $D$: $D=((\mathbf{x}^{(1)}, \mathbf{y}^{(1)}),$ $\dots$,
%$(\mathbf{x}^{(\eta)}, \mathbf{y}^{(\eta)}))$;
\STATE Store $\{\mathbf{x}^{(1)}, \cdots, \mathbf{x}^{(\eta)}\}$ and $( \mathbf{y}^{(1)},\dots, \mathbf{y}^{(\eta)})$ in $D$: $D=((\mathbf{x}^{(1)}, \mathbf{y}^{(1)}),$ $\dots$,
$(\mathbf{x}^{(\eta)}, \mathbf{y}^{(\eta)}))$;
\STATE Compute the non-dominated subset of $D$ and store it in $\pfa$;
\STATE $g=\eta$;
\WHILE{$g<=T_c$} 
	\STATE Train surrogate models $M$ based on $D$;
	\STATE Use an optimizer to find the promising point $\mathbf{x}^*$ based on surrogate models $M$, with the infill criterion $C$;
	\STATE Update $D$: $D = D \cup ((\mathbf{x}^*, y(\mathbf{x}^*)))$;
	\STATE Update $\pfa$ as a non-dominated subset of $D$;
	\STATE $g=g+1$;
\ENDWHILE
\STATE Return $\pfa$.
\end{algorithmic}
\label{alg:mobgo}
\end{algorithm}

First, a dataset $D$ is initialized and a Pareto-front approximation set $\pfa$ is computed, as shown in Algorithm \ref{alg:mobgo} from Step 1 to Step 5. The initialization of $D$ contains the generation of the decision vectors using Latin Hypercube Sampling method (LHS) \cite{lhs1979} (Step 1), calculation of the corresponding objective values (Step 2) and storage of this information in dataset $D$ (Step 3). This dataset $D$ will be utilized to build the Kriging models in the second part. 

The second part of MOBGO is the main loop, as shown in Algorithm \ref{alg:mobgo} from Step 6 to Step 12. This main loop starts with training Kriging models $M_i$ based on dataset $D$ (Step 7). Note that $M$ contains $d$ independent models for each objective function, and these models will be used as temporary objective functions instead of `true' objective functions in Step 8. Then, an optimizer finds a promising solution $\mathbf{x}^*$ by maximizing or minimizing an infill criterion $C$ (Step 8). Here, an infill criterion is calculated by its corresponding calculation formula, whose inputs include Kriging models $M$, the current Pareto-front approximation set $\pfa$, a target decision vector $\mathbf{x}^t$, etc. Theoretically, any single-objective optimization algorithm can be utilized as an optimizer to search for a promising solution $\mathbf{x}^*$. In this paper, the BI-population CMA-ES is chosen for its favorable performance on BBOB function testbed \cite{hansen2009benchmarking}. Step 9 and Step 10 will update the dataset $D$ by adding $(\mathbf{x}^*,y(\mathbf{x}^*))$ into $D$ and update the Pareto-front approximation set $\pfa$. The main loop from Step 6 to Step 12 will continue until $g$ meets the termination criterion $T_c$. The last part of MOBGO returns Pareto-front approximation set $\pfa$.

The choice of infill criterion $C$ at Step 8 distinguishes different types of MOBGO based algorithms. In this paper, EHVI-MOBGO and PoI-MOBGO, which set EHVI and PoI \cite{kushner1964poi,ulmer2003evolution,keane2006statistical} as the infill criterion C respectively, are compared in Section \ref{subsec:benchmark}.

\section[title]{Definitions\footnote{For the convenience of visualization and consistency, this paper only considers maximization problems. Minimization problems can always be re-written as maximization problems by multiplying the corresponding objective functions by $(-1)$.}}\label{sec:NotitionDefinition}
\emph{Pareto dominance}, or briefly \emph{dominance}, is a fundamental concept in MOO and provides an ordering relation on the set of potential solutions. \emph{Dominance} is defined as follows:
\begin{definition}[Dominance \textnormal{\cite{CoelloCoello2011}}]%
% $\forall \mathbf{y}, \mathbf{z} \in \mathbb{R}^d: \mathbf{y} \prec \mathbf{z} \Leftrightarrow (\forall i\in\{1, \dots, d\}: y_i \geq z_i) \mbox{ and }   \mathbf{y} \neq \mathbf{z}$ for maximization problems.
Given two decision vectors $\mathbf{x}^{(1)},\mathbf{x}^{(2)} \in \mathbb{R}^m $ and their corresponding objective values $\mathbf{y}^{(1)}=y(\mathbf{x}^{(1)})$, $\mathbf{y}^{(2)}=y(\mathbf{x}^{(2)})$ in a maximization problem, it is said that $\mathbf{y}^{(1)}$ dominates $\mathbf{y}^{(2)}$, being represented by $\mathbf{y}^{(1)} \prec \mathbf{y}^{(2)}$, iff \enskip $\forall i \in \{ 1, 2, \cdots, d \}: {y}_i(\mathbf{x}^{(1)}) \geq {y}_i(\mathbf{x}^{(2)})$ and $\exists j \in \{ 1, 2, \cdots, d \}: {y}_j(\mathbf{x}^{(1)}) > {y}_j(\mathbf{x}^{(2)})$.
\label{def:dominance}
\end{definition}

From the perspectives of searching and optimization, non-dominated points are of greater interest. 
%, because a point in a non-dominated space means a potential improvement of the objective function values.
The concept of \emph{non-dominance} is defined as: 
\begin{definition}[Non-dominance \textnormal{\cite{Michael2016book}}]
Given a decision vector set $\mathbf{X} \in \mathbb{R}^m$, and the image of the vector set  $\mathbf{Y} = \{y(\mathbf{x}) | \mathbf{x} \in \mathbf{X} \}$, the non-dominated subset of $\mathbf{Y}$ is defined as:
\begin{align*}
	nd(\mathbf{Y}) := \{ \mathbf{y} \in \mathbf{Y} | \nexists \mathbf{z} \in \mathbf{Y}: \mathbf{z} \prec \mathbf{y} \}  \numberthis
\end{align*}
\label{def:non-dominance}
\end{definition}
%\vspace{-0.5cm}
A vector $\mathbf{y} \in nd(\mathbf{Y})$ is called a non-dominated point of $\mathbf{Y}$. 
%A non-dominated set means that there is no solution better or equally good in all components of the objective space. 
%However, there could be solutions that are, at least, better in some component(s) with sacrificing the performance in the other component(s). The goal of MOO is trying to find all non-dominated solutions in a whole feasible search space, which is called \textbf{Pareto front}, 
%\begin{definition}[Dominated Space]
%Given a Pareto-front approximation set $\pfa$, a dominated space of $\pfa$ is a subset of an objective space $\mathbb{R}^d$ and is defined as: 
%\begin{align*}
%\mbox{dom}(\pfa) := \{ \mathbf{z} \in \mathbb{R}^d | \nexists \mathbf{z} \prec \mathbf{y}, \mathbf{y} \in \pfa \} \numberthis
%\end{align*}
%\label{def:dominated_space}
%\end{definition}
\begin{definition}[Dominated Subspace of a Set]
%Given a Pareto-front approximation set $\pfa$, a dominated space of $\pfa$ is a subset of an objective space $\mathbb{R}^d$ and is defined as: 
Let $\mathcal{P}$ be a subset of $\mathbb{R}^d$. The dominated subspace of $\mathcal{P}$ in $\mathbb{R}^d$, notation $\mbox{ dom }(\mathcal{P})$, is then defined as: 
\begin{align*}
\mbox{dom}(\pfa):= \{\, \mathbf{y} \in \mathbb{R}^d\, |\, \exists \mathbf{p} \in \mathcal{P} \mbox{ with } \mathbf{p} \prec \mathbf{y}\,  \} \numberthis
\end{align*}
\label{def:dominated_space}
\end{definition}
%\vspace{-1cm}

%CH -- below, change "HV does not require to know the Pareto front" to "HV does not require knowing the Pareto front"

\begin{definition}[Non-Dominated Space of a Set]
Let $\mathcal{P}$ be a subset of $\mathbb{R}^d$ and let $\mathbf{r} \in \mathbb{R}^d$ be such that $\forall \mathbf{p} \in \mathcal{P}: \mathbf{p} \prec \mathbf{r}$. The non-dominated space of $\mathcal{P}$ with respect to $\mathbf{r}$, denoted as $\mbox{ndom}(\pfa)$, is then defined as:
\begin{align*}
\mbox{ndom} (\pfa):= \{ \mathbf{y} \in \mathbb{R}^d\, |\, \mathbf{y} \prec \mathbf{r} \mbox{ and } \not \exists \mathbf{p} \in \mathcal{P} \mbox{ such that } \mathbf{p} \prec \mathbf{y} \, \} \numberthis
\end{align*} 
\end{definition}
Note that the notion of dominated space as well as the notion of non-dominated space of a set can also be defined for (countably and non-countably) infinite sets $\mathcal{P}$.

The \emph{Hypervolume Indicator} (HV), introduced in \cite{zitzler1999multiobjective}, is one of the essential unary indicators for evaluating the quality of a Pareto-front approximation set. Its theoretical properties are discussed in \cite{zitzler2003performance}. Notably, HV does not require the knowledge of the Pareto front in advance. The maximization of HV leads to a high-qualified and diverse Pareto-front approximation set.
The \emph{Hypervolume Indicator} is defined as follows:
\begin{definition}[Hypervolume Indicator]
Given a finite approximation to a Pareto front, say $\pfa = \{ \mathbf{y}^{(1)}, \dots, \mathbf{y}^{(n)}\} \subset \mathbb{R}^d$, the \emph{Hypervolume Indicator} of $\pfa$ is defined as the $d$-dimensional Lebesgue measure of the subspace dominated by $\pfa$ and bounded below by a reference point $\mathbf{r}$:
\begin{align*}
\mbox{HV}(\pfa) = \lambda_d(\cup_{\mathbf{y} \in \pfa} [\mathbf{r}, \mathbf{y}]) \numberthis
\end{align*}
with $\lambda_d$ being the Lebesgue measure on $\mathbb{R}^d$.
\label{def:HI}
\end{definition}
The hypervolume indicator measures the size of the dominated subspace
bounded below by a reference point $\mathbf{r}$. This reference point needs to be provided by users. Theoretically, in order to get the extreme non-dominated points, this reference point should be chosen in a way that it is dominated by all elements of a Pareto-front approximation set $\pfa$ during the optimization process. However, there is no requirement of setting the reference point in practice if the user is not interested in extreme non-dominated points.  

Another important infill criterion is \emph{Hypervolume Improvement}, which is also called the \emph{Improvement of Hypervolume} in \cite{emmerich2008computation}.
 %and \emph{the most likely improvement} (MLI) in \cite{emmerich2005emo}. 
%The basic idea of HVI is the HV change of a Pareto front approximation set $\pfa$  before and after adding an evaluated point $\mathbf{y}$ in it. 
The definition of \emph{Hypervolume Improvement} is:

\begin{definition}[Hypervolume Improvement]
Given a finite collection of vectors $\pfa \subset \mathbb{R}^d$, the \emph{Hypervolume Improvement} of a vector $\mathbf{y} \in \mathbb{R}^d$ is defined as:
\begin{align*}
\label{eq:hvi}
\mbox{HVI}(\mathbf{y}, \pfa ) = \mbox{HV}(\pfa \cup \{\mathbf{y}\}) - \mbox{HV}(\pfa) \numberthis
\end{align*}
When we want to emphasize the reference point $\mathbf{r}$, the notation  $\mbox{HVI}(\mathbf{y}, \pfa,\mathbf{r})$ will be used to denote \emph{Hypervolume Improvement}.
\end{definition}

\begin{example}
Figure \ref{fig:ExampleHVI} illustrates the concept of \emph{Hypervolume Improvement} using two examples. The first example, on the left, is a 2-D example: Suppose a Pareto-front approximation set is $\pfa$, which is composed by $\mathbf{y}^{(1)}= (1,2.5)^\top$, $\mathbf{y}^{(2)}= (2,1.5)^\top$ and $\mathbf{y}^{(3)} = (3,1)^\top$. When a new point $\mathbf{y^{(+)}} = (2.8,2.3)^\top$ is added, the \emph{Hypervolume Improvement} $HVI(\mathbf{y^{(+)}},\pfa, \mathbf{r})$ is the area of the yellow polygon. The second example (on the right in Figure 1) illustrates the \emph{Hypervolume Improvement} by means of a 3-D example. Assume a Pareto-front approximation set is 
$\pfa =$(
$\mathbf{y}^{(1)}=(4,4,1)^\top$, 
$\mathbf{y}^{(2)}=(1,2,4)^\top$, 
$\mathbf{y}^{(3)}=(2,1,3)^\top$). 
The \emph{Hypervolume Improvement} of $\mathbf{y}^{(+)}=(3,3,2)^\top$ relative to $\pfa$ is given by the joint volume covered by the yellow slices.
\begin{figure}[!ht]
\centering
\includegraphics[width=0.45\textwidth]{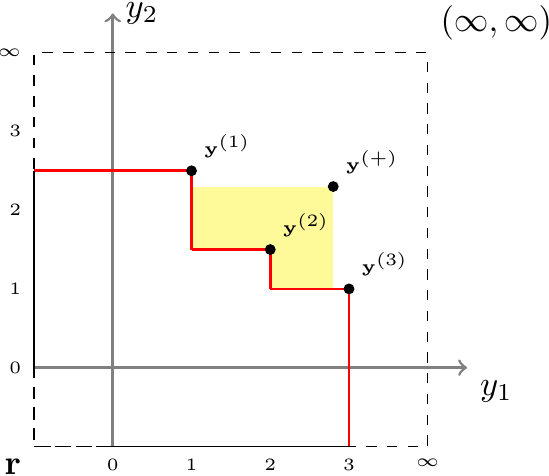}
\includegraphics[width=0.45\textwidth]{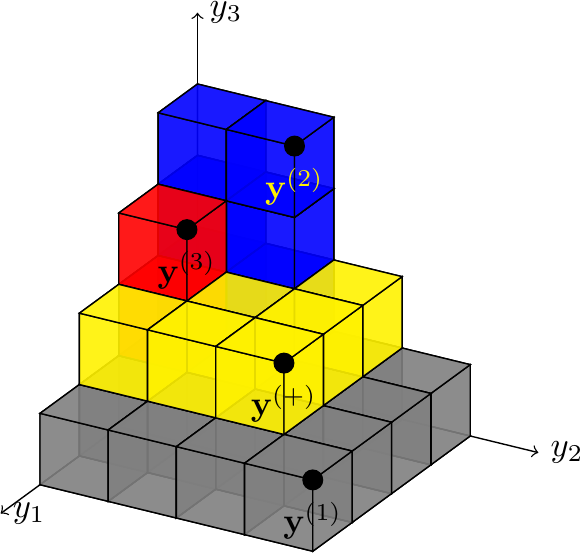}
\caption{\label{fig:ExampleHVI} The left and right figures illustrate \emph{Hypervolume Improvement} in a 2-D and a 3-D example, respectively.}
\end{figure}
\end{example}

\emph{Probability of Improvement} (PoI) is an important criterion in MOBGO. It was first introduced by Stuckman in \cite{stuckman1988global}. Later, Emmerich et al. \cite{emmerich2006single} generalized it to multi-objective optimization. PoI is defined as:
\begin{definition}[Probability of Improvement]
Given parameters of the multivariate predictive distribution $\boldsymbol\mu$, $\boldsymbol\sigma$ and the Pareto-front approximation set $\pfa$, the \emph{Probability of Improvement} is defined as:
\begin{align*}
\mbox{PoI}(\boldsymbol\mu, \boldsymbol\sigma, \pfa) :&= \int_{\mathbb{R}^d} \mathrm{I}(\mathbf{y} \mbox{ impr } \pfa )  \boldsymbol{\pdf}_{\boldsymbol\sigma, \boldsymbol\mu}(\mathbf{y}) d\mathbf{y} \qquad \qquad
\mathrm{I}(v) =
\left\{ 
\begin{array}{ll} 
1 & \mbox{ if } v= \mathrm{true}\\
0 & \mbox{ if } v= \mathrm{false}
\end{array} 
\right.
\numberthis
\label{def:poi}
\end{align*}
where $\boldsymbol{\pdf}_{\boldsymbol{\mu}, \boldsymbol{\sigma}}$ is a multivariate independent normal distribution with the mean values $\boldsymbol{\mu} \in \mathbb{R}^d$ and the standard deviations $\boldsymbol{\sigma} \in \mathbb{R}^d_+$. Here, ($\mathbf{y} \mbox{ impr } \pfa$) represents $\mathbf{y} \in \mathbb{R}^d$ as an improvement with respect to $\pfa$, if and only if the following holds: $\mathbf{y} \prec \mathbf{r}$ and $\forall \mathbf{p} \in \mathcal{P}: \neg (\mathbf{p} \prec \mathbf{y})$. 
\end{definition}
%
%The following definition formalizes the notion of a point $\mathbf{y}$ being an improvement with respect to a set $\mathcal{P}$. 
%\begin{definition}[asdf ]
%Let $\mathbf{r} \in \mathbb{R}^d$ be the reference point and $\mathcal{P} \subset \mathbb{R}^d$ be such that $r \prec p$ for all $p \in \mathcal{P}$. Then $\mathbf{y} \in \mathbb{R}^d$ is an improvement with respect to $\mathcal{P}$ and $\mathbf{r}$, notation $\mathbf{y} \mbox{ impr } \mathcal{P}$, if and only if the following holds:
%\begin{itemize}
%\item $\mathbf{y} \prec \mathbf{r}$, and
%\item $\forall p \in \mathcal{P}: \neg (\mathbf{p} \prec \mathbf{y}) $
%\end{itemize}
%\end{definition}

In Equation (\ref{def:poi}), $\mbox{I}(\mathbf{y} \mbox{ impr } \pfa)=1$ means that $\mathbf{y}$ is an element of the non-dominated space of $\mathcal{P}$. In other words, $\mathbf{y} \in [\mathbf{r}, \infty^d] \setminus \mbox{ dom }(\mathcal{P})$ if $\mbox{I}(\mathbf{y} \mbox{ impr } \pfa)=1$.
A reference point $\mathbf{r}$ is not indicated in Equation (\ref{def:poi}) because $\mathbf{r}$ must be chosen as $[-\infty]^d$ in PoI. Therefore, PoI is a reference-free infill criterion.
%Moreover, a reference point $\mathbf{r}$ in the definition of the PoI can be chosen as $[-\infty]^d$, while the \emph{Hypervolume} based infill criterion (e.g. EHVI, HVI) can not. Therefore, PoI is a reference-free infill criterion. 

\begin{definition}[Expected Hypervolume Improvement]
Given parameters of the multivariate predictive distribution $\boldsymbol\mu$, $\boldsymbol\sigma$ and the Pareto-front approximation set $\pfa$, the \emph{expected hypervolume improvement} is defined as:
\begin{align*}
EHVI(\boldsymbol\mu, \boldsymbol\sigma, \pfa, \mathbf{r}) := \int_{\mathbb{R}^d} \mbox{HVI}(\pfa, \mathbf{y}, \mathbf{r}) \cdot \boldsymbol{\pdf}_{\boldsymbol\sigma, \boldsymbol\mu}(\mathbf{y}) d\mathbf{y}  \numberthis
\label{def:EHVI}
\end{align*}
%where  $\boldsymbol{\pdf}_{\boldsymbol{\mu}, \boldsymbol{\sigma}}$ is the multivariate independent normal distribution with the mean values $\boldsymbol{\mu} \in \mathbb{R}^d$,  and the standard deviations $\boldsymbol{\sigma} \in \mathbb{R}^d_+$.
\end{definition}

\begin{figure}[ht]
\centering
\includegraphics[width=0.9\textwidth]{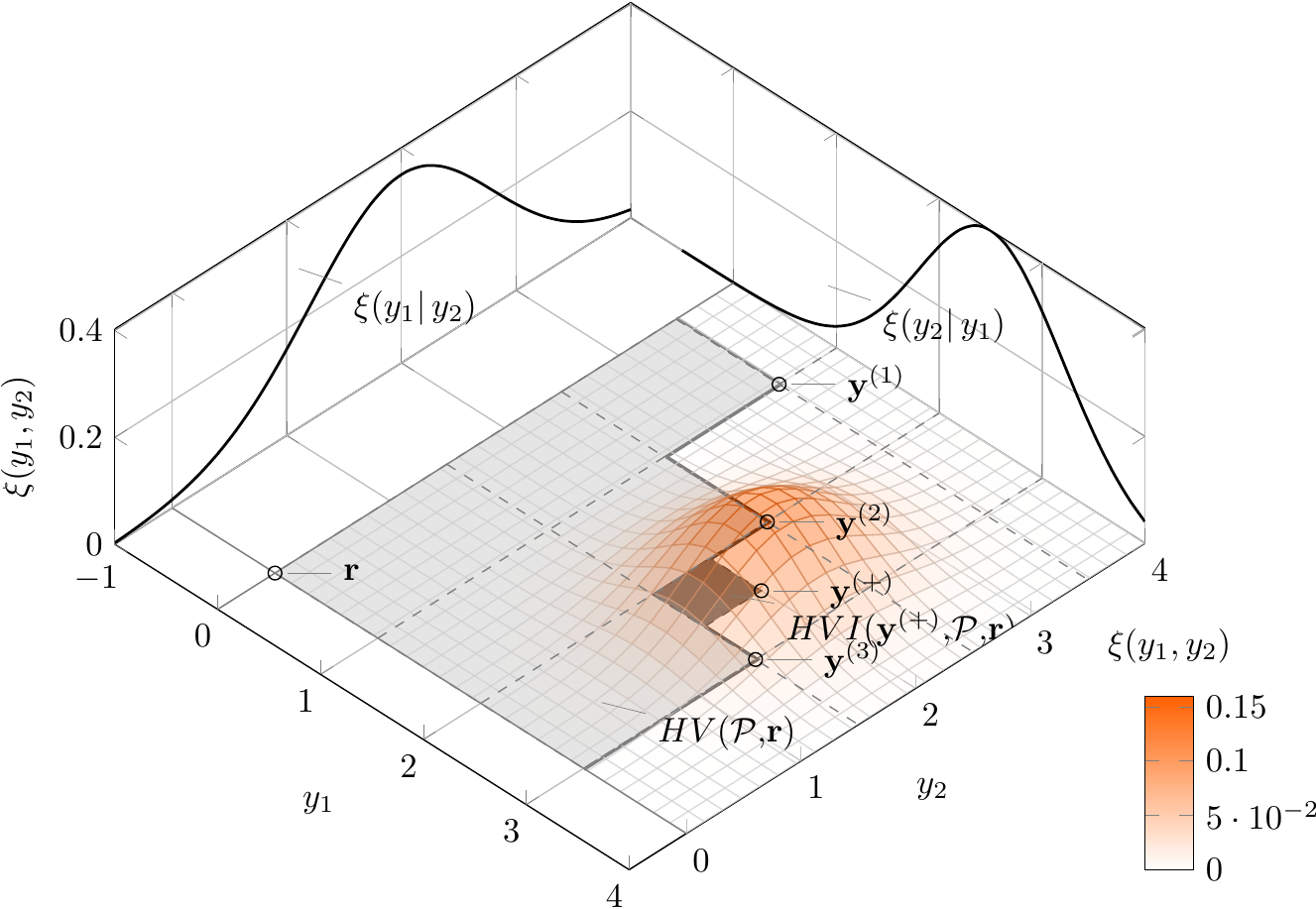}
\caption{\label{fig:ehvi2dexample}Expected hypervolume improvement in 2-D (cf. Example \ref{ex:ehvi}).}
\end{figure}

\begin{example}
\label{ex:ehvi}
An illustration of the 2-D EHVI is shown in  Figure \ref{fig:ehvi2dexample}. 
The light gray area is the dominated subspace of $\pfa = \{{\mathbf{y}^{(1)}} = (3,1)^\top,$ $\mathbf{y}^{(2)}=(2, 1.5)^\top,$ $ {\mathbf{y}^{(3)}} = (1, 2.5)^\top\}$ bounded by the reference point $\mathbf{r} = (0,0)$. The bivariate Gaussian distribution has the parameters
$\mu_1=2.5,$ $\mu_2=2,$ $\sigma_1 = 0.7,$ $\sigma_2 = 0.8$.
The probability density function ($\boldsymbol\pdf$) of the bivariate Gaussian distribution is indicated as a 3-D plot. Here $\mathbf{y}^{(+)}$ is a sample from this distribution and the area of improvement relative to $\pfa$ is indicated by the dark shaded area. Variables $y_1$ and $y_2$ stand for the first and the second objective values, respectively.
\end{example}

For computing integrals of EHVI in Section \ref{sec:MCC}, it is useful to define $\Delta$ and $\Psi_{\infty}$ functions.
\begin{definition}[$\Delta$ function \textnormal{(see also \cite{Michael2016book,yang2017phdthesis})}]
For a given vector of objective function values $\mathbf{y} \in \mathbb{R}^d$ and $\mathbf{y} \not\in \pfa$, $\Delta(\mathbf{y}, \pfa, \mathrm{r})$ is the subset of the vectors in $\mathbb{R}^d$ which are exclusively dominated by the vector $\mathbf{y}$ but not by elements in $\pfa$, and which dominate the reference point $\mathbf{r}$, that is:
\begin{equation}
\Delta(\mathbf{y}, \pfa, \mathbf{r}) = \{\mathbf{z} \in \mathbb{R}^d \ | \  \mathbf{y} \prec \mathbf{z} \mbox{ and }    \mathbf{z} \prec \mathbf{r} \mbox{ and } \not\exists \mathbf{q} \in \pfa:  \mathbf{q} \prec  \mathbf{z}\}
\end{equation}
\end{definition}

\begin{definition}[$\Psi_{\infty}$ function \textnormal{(see also \cite{hupkens2015faster})}]
\label{def:gaussian-psi}
Let $\phi(s)= 1/\sqrt{2\pi}e^{-\frac{1}{2}s^2} (s\in \mathbb{R})$ denote the PDF ($\pdf$) of the standard normal distribution. Moreover, let $\Phi(s)= \frac{1}{2}\left(1 + \erf \left(\frac{s}{\sqrt{2}}\right)\right)$ denote its cumulative probability distribution function (CDF), and $\erf$ denote the Gaussian error function. The general normal distribution with mean $\mu$ and standard deviation $\sigma$ has PDF $\pdf_{\mu,\sigma}(s)=\phi_{\mu, \sigma}(s) = \frac{1}{\sigma}\phi(\frac{s-\mu}{\sigma})$  and its CDF is $\Phi_{\mu, \sigma}(s) = \Phi(\frac{s-\mu}{\sigma})$. Then the function $\Psi_{\infty}(a,b,\mu,\sigma)$ is defined as:
\begin{align*}
\Psi_{\infty}(a,b,\mu,\sigma)
:&= \int_{b}^{\infty}(z-a)\dfrac{1}{\sigma}\phi \left(\dfrac{z-\mu}{\sigma}\right)dz \label{eq:exipsi_inf}
  \numberthis
\end{align*}
\end{definition}
One can easily show that $\psi_{\infty}(a,b,\mu,\sigma) =\sigma\phi\left(\dfrac{b-\mu}{\sigma}\right) + (\mu-a)\left[1-\Phi\left(\dfrac{b-\mu}{\sigma}\right)\right]$.
\section[Section title]{Efficient EHVI Calculation}\label{sec:MCC}
This section mainly discusses an efficient partitioning method for a non-dominated space and how to employ this partitioning method to calculate EHVI and PoI.  
\subsection{Partitioning a non-dominated space}\label{subsec:Partitioning}	
The efficiency of an infill criterion calculation is determined by a non-dominated search algorithm and the number of integration slices. The main idea of the partitioning method is to separate the integration volume (a non-dominated space) into as few integration slices as possible.
Then, the integral of the criterion is calculated within each integration slice. The value of the criterion is the sum of its contribution in every integration slice.
\subsubsection{The 2-D case}
%In the 2-D case, the partitioning method is simple and has already been published by Emmerich et al. \cite{Michael2016book}. It uses a new way to integrate the EHVI that does not require a partitioning into a grid, as it was neccessary for previously proposed algorithms by Hupkens et al. \cite{hupkens2015faster} and Couckuyt et al. \cite{couckuyt2014fast}, but can make use of arbitrary box partitionings. For the sake of completeness and to introduce this integration technique, we will introduce it here briefly. 
In the 2-D case, the partitioning method is simple and has already been published by Emmerich et al. \cite{Michael2016book}. 
Given a Pareto-front approximation set $\pfa$ with $n$ elements, the algorithm in \cite{Michael2016book} adopts a new way to derive the EHVI calculation formulas and only partitions a non-dominated space into $n+1$ integration slices, instead of $(n+1)^2$ grids in \cite{hupkens2015faster, couckuyt2014fast}. For the sake of completeness, we will introduce this integration technique here briefly. 

\begin{figure}[ht]
\centering
\includegraphics[width=0.49\textwidth]{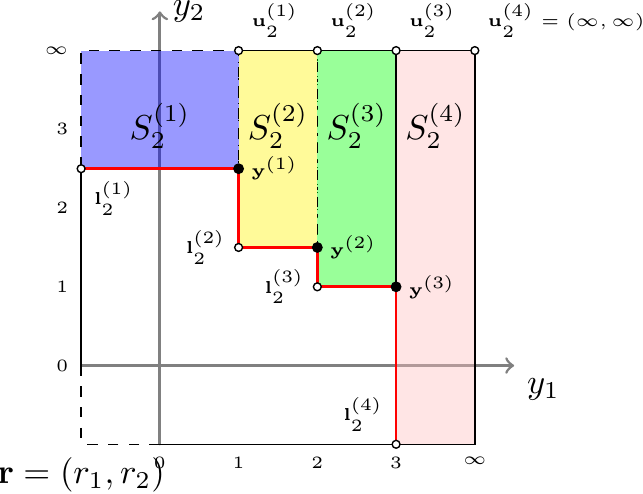} 
\caption{\label{fig:Partioning2D} Partitioning of the 2-D integration region into slices.}
\end{figure}

Suppose $\mathbf{Y}={\mathbf{y}^{(1)}, \dots, \mathbf{y}^{(n)}}$ and $d=2$, an integration space (a non-dominated space) of $\mathbf{Y}$ can be divided into $n+1$ disjoint integration slices ($S_2^{(i)}, i=1,\dots,n+1$) by drawing lines parallel to $y_2$-axis at each element in $\mathbf{y}$, as indicated in Figure \ref{fig:Partioning2D}. Then, each integration slice can be expressed by its lower bound ($\mathbf{l}_2^{(i)}$) and upper bound ($\mathbf{u}_2^{(i)}$). 
%In order to define the slices formally, augment $\pfa$ with two sentinels: $\mathbf{y}^{(0)} = (r_1, \infty)^\top$ and $\mathbf{y}^{(n+1)} = (\infty, r_2)^\top$. 
In order to define the slices formally, we argue a Pareto-front approximation set $\pfa$ with two sentinels: $\mathbf{y}^{(0)} = (r_1, \infty)^\top$ and $\mathbf{y}^{(n+1)} = (\infty, r_2)^\top$. 
Then, the integration slices for the 2-D case are defined by: 
\begin{align*}
S_2^{(i)} = (\mathbf{l}_2^{(i)},\mathbf{u}_2^{(i)}) = 
\left(\left(\begin{array}{c}l_1^{(i)}  	\\ 	l_2^{(i)} 	\end{array} 	\right), 
\left(\begin{array}{c}u_1^{(i)} 		\\ 	u_2^{(i)} 	\end{array}\right) \right)   = 
\left(\left(\begin{array}{c}y_1^{(i-1)} \\ 	y_2^{(i)} 	\end{array} 	\right), 
\left(\begin{array}{c}y_1^{(i)} 		\\ 	\infty    	\end{array}\right) \right), 
i=1, \dots, N_2 \numberthis 
\end{align*}\label{def:slice2d}

%In the 2-D case, it is straightforward to calculate the number of integration slices, namely, $N_2=n+1$.
In the 2-D case, the number of integration slices is straightforward, namely, $N_2=n+1$.
\subsubsection{The 3-D case}
\ifdefined\DEBUG
    DEBUG was on
\else
\begin{figure}[ht]
	\centerline{\includegraphics[width=0.49\textwidth]{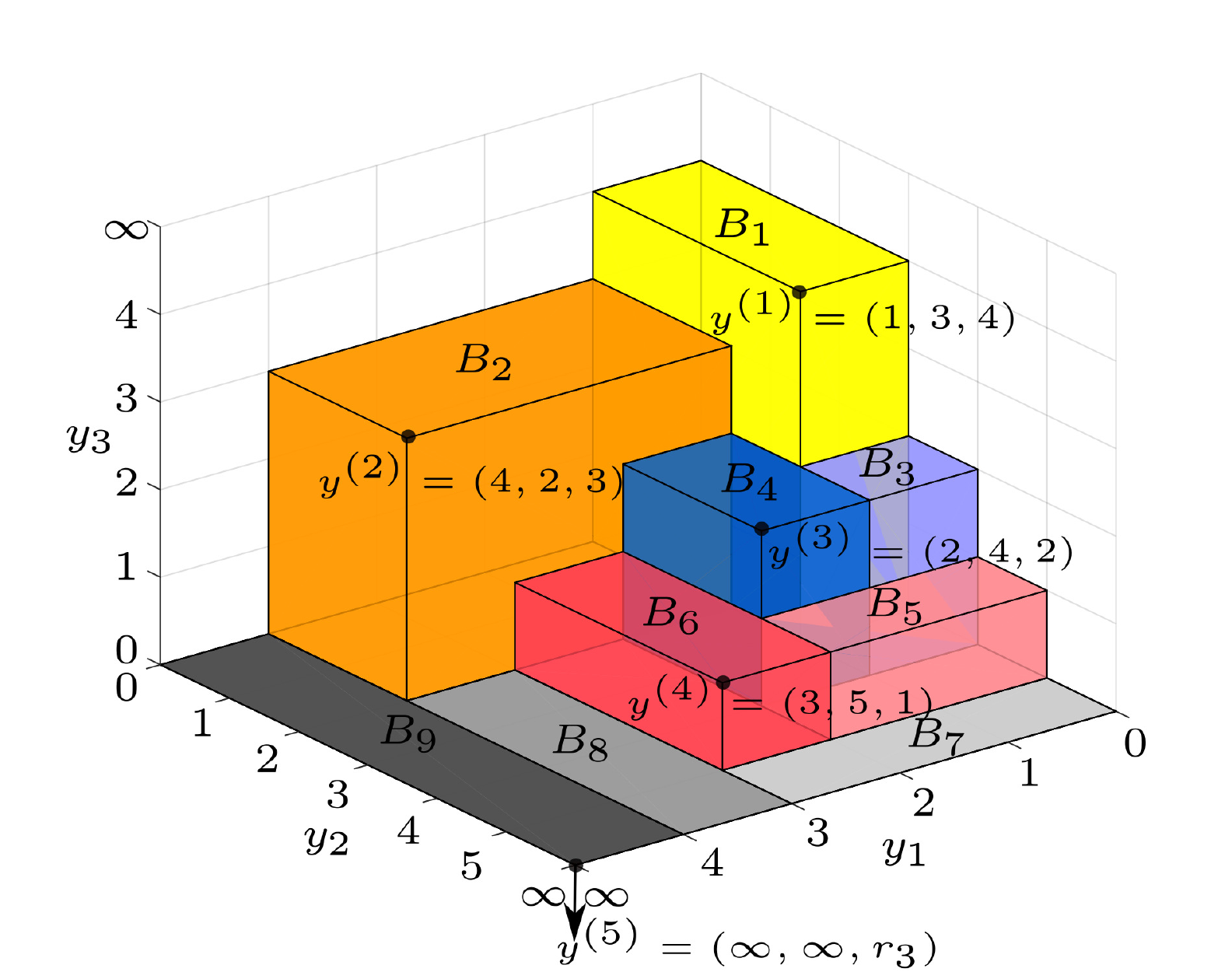}
	\includegraphics[width=0.49\textwidth]{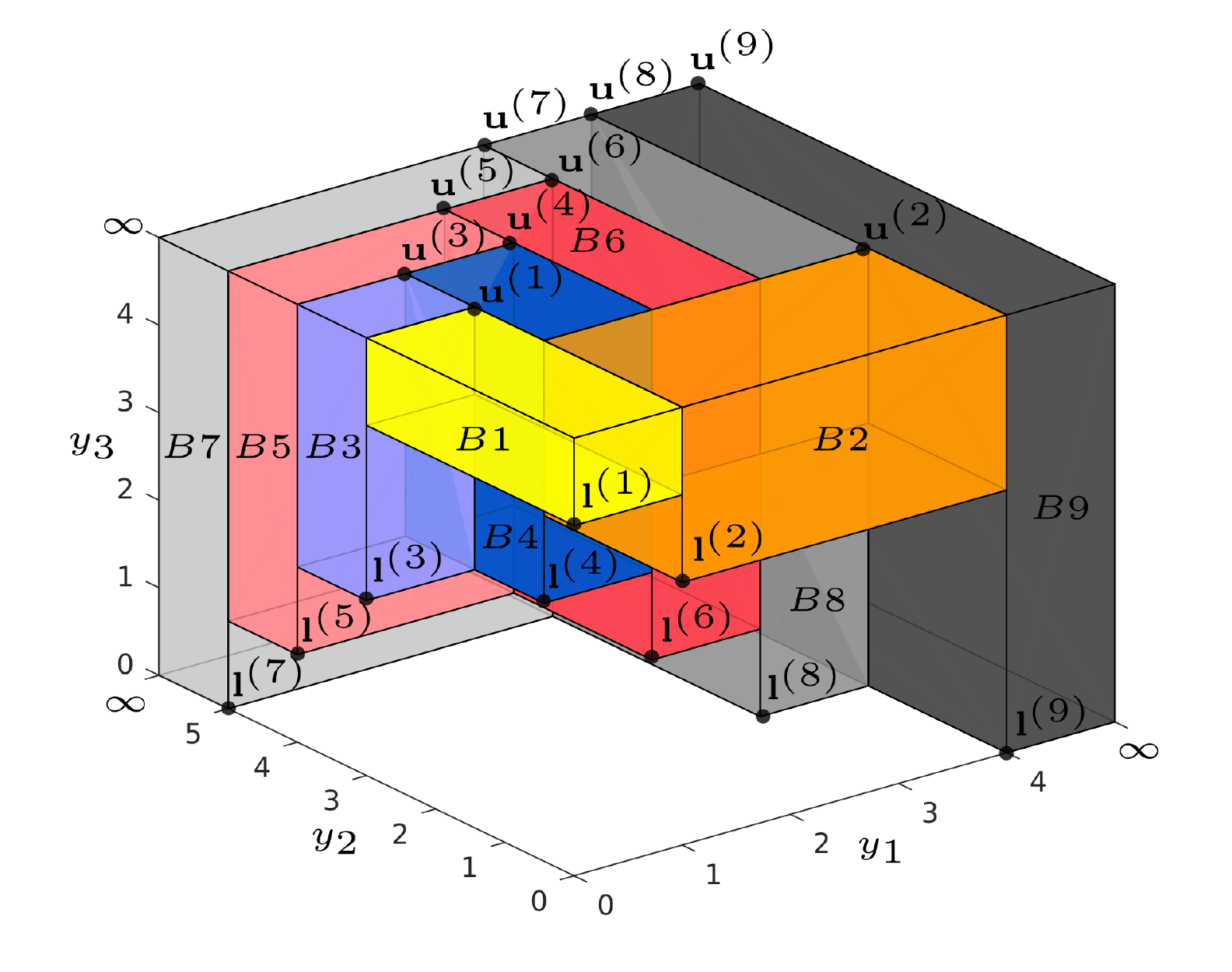} }
	\begin{center}
	\includegraphics[width=0.49\textwidth]{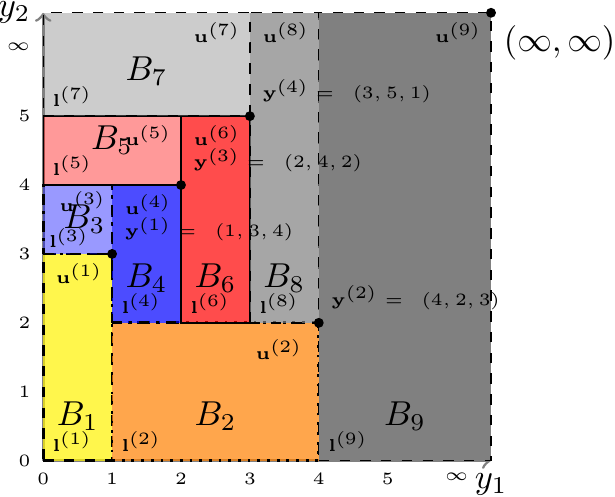} 
	\end{center}
    \caption{Upper left: 3-D Pareto-front approximation. Upper right: Integration slices in 3-D. Lower center: The projection of 3-D integration slices into the $y_1y_2$-plane, each slice can be described by lower bound and upper bound.}
    \label{fig:box_3d}
\end{figure}
\fi
Similar to the 2-D partitioning method, in the 3-D case, each integration slice can be defined by its lower bound ($\mathbf{l}_3$) and upper bound ($\mathbf{u}_3$). Since the upper bound of each integration slice is always $\infty$ in the $y_3$ axis, we can describe each integration slice as follows:
\begin{align*}
S_3^{(i)} = (\mathbf{l}_3^{(i)}, \mathbf{u}_3^{(i)})=\left(\left(\begin{array}{c}l_1^{(i)}\\l_2^{(i)}\\ l_3^{(i)}\end{array}\right), \left(\begin{array}{c}u_1^{(i)}\\u_2^{(i)}\\ \infty\end{array}\right) \right), \quad \quad i=1, \dots, N_3 \numberthis 
\end{align*}

\begin{example}
An illustration of integration slices is shown in Figure \ref{fig:box_3d}. A Pareto front set $\pfa$ is composed by 4 points ($\mathbf{y}^{(1)}=(1,3,4)^\top, \mathbf{y}^{(2)}=(4,2,3)^\top, \mathbf{y}^{(3)}=(2,4,2)^\top$ and $\mathbf{y}^{(4)}=(3,5,1)^\top$), and this Pareto front is shown in the upper left figure. The upper right figure represents the partitioned integration slices of $\pfa$. 
The lower center figure illustrates the projection of the upper right figure onto the $y_1y_2$-plane with rectangle slices and $\mathbf{l}, \mathbf{u}$. The rectangular slices, which share a similar color but of different opacity, represent integration slices with the same value of $y_3$ in their lower bound. 
The lower bound of the 3-D integration slice $B_4$ is $\mathbf{l}_3^{(4)} = (1,2,2)^\top$, and the upper bound of the slice is $\mathbf{u}_3^{(4)} = (2,4,\infty)^\top$.
\end{example}

\begin{algorithm}[!ht]
\KwIn{($\mathbf{y}^{(1)}, \cdots,\mathbf{y}^{(n)}$): mutually non-dominated $\mathbb{R}^3$-points sorted by third coordinate ($y_3$) in descending order}
\KwOut{$S_3^{(1)}, \cdots, S_3^{(i)}, \cdots, S_3^{(N_3)}$}
\nl $\mathbf{y}^{(n+1)}=(\infty, \infty, r_3)$ \;
\nl Initialize AVL tree T for 3-D points\\
    Insert $\mathbf{y}^{(1)}$, $(\infty, r_2, \infty)^{\top}$ and $(r_1, \infty, \infty)^{\top}$ into T\;
\nl Initialize the number of integration slices $n_b=1$\;
\nl Initialize $EHVI=0$\;
\nl \For(\tcc*[f]{Main loop}){$i=2$ \KwTo $n+1$}{
\nl Retrieve the following information from tree T:\\
\nl \quad r: index of the successor of $\mathbf{y^{(i)}}$ in $y_1$-coordinate (right neighbor);\\
\nl \quad t: index of the successor of $\mathbf{y^{(i)}}$ in $y_2$-coordinate (left neighbor);\\
\nl \quad d[1], $\cdots$, d[s]: indices of points dominated by $\mathbf{y^{(i)}}$ in $y_1y_2$-plane, sorted ascendingly in the first coordinate($y_1$)\;

\nl	$S^{(n_b)}_3.l_3=y^{(i)}_3$, \quad$S^{(n_b)}_3.u_2=y^{(i)}_2$, \quad$S^{(n_b)}_3.u_3=\infty$ \;
\nl \If(\tcc*[f]{Case 1}){$s==0$}{\nl $S^{(n_b)}_3.l_1=y^{(t)}_1$, \quad $S^{(n_b)}_3.l_2=y^{(r)}_2$, \quad$S^{(n_b)}_3.u_1=y^{(i)}_1$\;
%\nl $EHVI=EHVI+$ calculation\_3d$(\boldsymbol{\mu}, \boldsymbol{\sigma}, B_{n_b})$ \;
\nl	$n_b=n_b+1$ \;	
	}
\nl \Else(\tcc*[f]{Case 2}){
\nl \For{$j=1$ \KwTo $s+1$}{

\nl	\If{$j==1$}{\nl $S^{(n_b)}_3.l_1=y^{(t)}_1$, \quad $S^{(n_b)}_3.l_2=y^{\text{(d[1])}}_2$,\quad $S^{(n_b)}_3.u_1=y^{(d[1])}_1$\;}
\nl	\ElseIf{$j==s+1$}{\nl $S^{(n_b)}_3.l_1=y^{\text {(d[s])}}_1$, \quad$S^{(n_b)}_3.l_2=y^{(r)}_2$, \quad$S^{(n_b)}_3.u_1=y^{(i)}_1$\;}
\nl	\Else{\nl $S^{(n_b)}_3.l_1=y^{\text{(d[j-1])}}_1$, \quad$S^{(n_b)}_3.l_2=y^{\text{(d[j])}}_2$, \quad$S^{(n_b)}_3.u_1=y^{\text{(d[j])}}_1$\;}
% \nl $EHVI=EHVI+$ calculation\_3d$(\boldsymbol{\mu}, \boldsymbol{\sigma}, B_{n_b})$ \;
\nl	$n_b=n_b+1$ \;		
	}
	}
\nl Discard $\mathbf{y}^{\text {(d[1])}}, \cdots , \mathbf{y}^{\text {(d[s])}}$ from tree T\;
\nl Insert $\mathbf{y}^{(i)}$ in tree T.	
}
\nl Return $S^{(1)}_3, \cdots, S^{(i)}_3, \cdots, S^{(N_3)}_3$
\caption{{\bf Integration slice acquisition in 3-D case} \label{Algorithm}}
\end{algorithm}

\begin{figure}[!ht]
\centering
\includegraphics[scale=1]{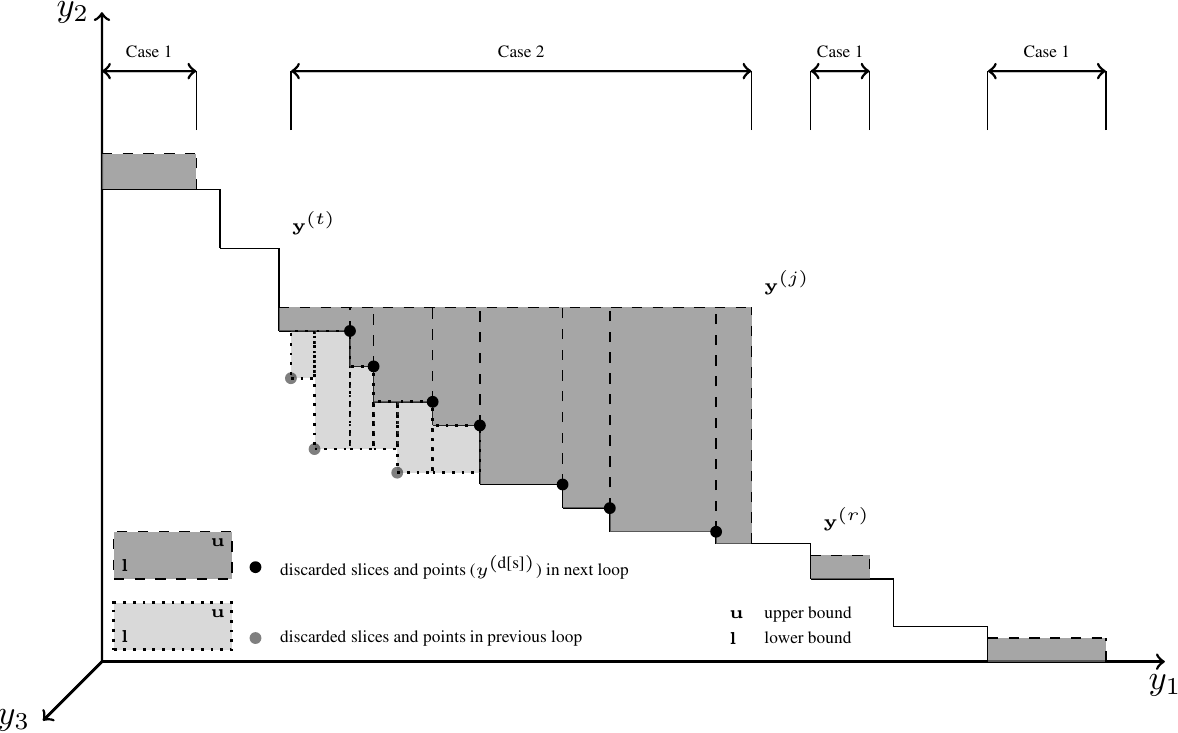}
\caption{Boundary search for slices in 3-D case.}
\label{fig:search_box}
\end{figure}

Algorithm \ref{Algorithm} describes how to obtain the slices $S_3^{(1)}$, $\dots$, $S_3^{(i)}$, $\dots$, $S_3^{(N_3)}$ with the corresponding lower and upper bounds ($\mathbf{l}_3^{(i)}$ and $\mathbf{u}_3^{(i)}$). The partitioning algorithm is similar to the sweep line algorithm described in \cite{emmerich2011computing}. The basic idea of this algorithm is to use an AVL tree to process points in descending order of the $y_3$ coordinate. For each such point, say $\mathbf{y}^{(i)}$, the algorithm finds all the points $(\mathbf{y}^{(d[1])},\dots, \mathbf{y}^{(d[s])})$ which are dominated by $\mathbf{y}^{(i)}$ in the $y_1y_2$-plane and inserts $\mathbf{y}^{(i)}$ into the tree. Moreover, because of $\mathbf{y}^{(i)}$, the algorithm will also discard all the points ($\mathbf{y}^{\text{(d[1])}}$, $\dots$, $\mathbf{y}^{\text{(d[s])}}$) from the AVL tree. See Figure \ref{fig:search_box} for describing one such iteration. In each iteration, $s+1$ slices are created by coordinates of the points $\mathbf{y}^{\text{(t)}}$, $\mathbf{y}^{\text{(d[1])}}$, $\dots$, $\mathbf{y}^{\text{(d[s])}}$, $\mathbf{y}^{(r)}$, and $\mathbf{y}^{(i)}$ as illustrated in Figure \ref{fig:search_box}.

The number of the integration slices in the 3-D case $N_3$ is $2n+1$ where all points are in general position (for each $i, i=1,\dots, d$: the $i$-th coordinate is different for each pair of points in $\mathbf{Y}$). Otherwise, $2n+1$ provides an upper bound for the obtained number of slices. 

\emph{Proof:} In the algorithm, each point $\mathbf{y}^{(i)}|_{i=1,\dots,n}$ creates two slices. The first one, say slice $A^{(i)}$, is created when the point $\mathbf{y}^{(i)}$ is \emph{added} to the AVL tree. Another slice, say slice $S_3^{(i)}$, is created when the point $\mathbf{y}^{(i)}$ is \emph{discarded} from the AVL tree due to domination by another point, say $\mathbf{y}^{(s)}$, in the $y_1y_2$-plane. 
%
%and a slice, say slice $S_3^{(i)}$, when it is discarded from the AVL tree due to domination by another point, say $\mathbf{y}^{(s)}$, in the $y_1y_2$-plane. 
%
%In the algorithm each point $\mathbf{y}^{(i)},i=1,\dots,n$ creates a slice, say slice $A^{(i)}$, when it is created and a slice, say slice $S_3^{(i)}$, when it is discarded from the AVL tree due to domination by another point, say $\mathbf{y}^{(s)}$, in the $y_1y_2$-plane.
These two slices are defined as follows $A^{(i)} = ((y^{(t)},y_2^{(l2)},y_3^{(i)})^\top,(y_1^{(u1)},y_2^{(i)},\infty)^\top)$ whereas $y_2^{(l2)}$ is either $y_2^{(r)}$ if no point is dominated by $\mathbf{y}^{(i)}$ in the $y_1y_2$-plane, or $y_2^{(d[1])}$, otherwise. Moreover, $S_3^{(i)}=((y^{(i)}_1,y^{(r)}_2,y^{(s)}_3)^\top,$ $(y^{(u)}_1,y^{(s)}_2,\infty)^\top)$ and $\mathbf{y}^{(u)}$ denote either the right neighbor among the newly dominated points in the $y_1y_2$-plane, or $\mathbf{y}^{(s)}$ if $\mathbf{y}^{(i)}$ is the rightmost point among all newly dominated points. In this way, each slice can be attributed to exactly one point in $\pfa$, except the slice that is created in the final iteration. In the final iteration, one additional point $\mathbf{y}^{(n+1)}=(\infty, \infty,\infty)^\top$ is added to the AVL tree. This point will create a new slice when it is added, but because it is never discarded, it adds only a single slice. Therefore, $2n + 1$ slices are created in total.

\subsubsection{Higher dimensional cases}\label{subsec:ehvi_hd}
In higher dimensional cases, the non-dominated space can be partitioned into axis aligned hyperboxes, similar to the 3-D case. In the $d$-dimensional case ($d\geq 4$), the hyperboxes can be denoted by $S_d^{(1)}, \dots, S_d^{(i)}, \dots, S_d^{N_d}$ with their lower bounds ($\mathbf{l}^{(1)}, \dots, \mathbf{l}^{(N_d)}$) and upper bounds ($\mathbf{u}^{(1)}, \dots, $ $\mathbf{u}^{(N_d)}$). Here, $N_d$ is the number of hyperboxes and has the same definition as $N_2$ and $N_3$. The hyper-integral box $S_d^{(i)}$ is defined as: 
\begin{align*}
S_d^{(i)} 
= (\mathbf{l}_d^{(i)}, \mathbf{u}_d^{(i)})
= \big( (l_1^{(i)}, \cdots, l_d^{(i)})^{\top}, (u_1^{(i)}, \cdots, \infty)^{\top} \big) \quad \quad i=1, \dots, N_d \numberthis
\end{align*}

An efficient algorithm for partitioning a higher dimensional, non-dominated space is proposed in this section, which is based on two state-of-the-art algorithms DKLV17 \cite{dachert2017efficient} by D{\"a}chert et al. and LKF17 \cite{lacour2017box} by Lacour et al. Here, algorithm DKLV17 is an efficient algorithm to locate the local lower bound points\footnote{For the definition of the local lower bound points $\mathbf{L}_d$, see \cite{dachert2017efficient}.} ($\mathbf{L}_d$) in a dominated space for maximization problems, based on a specific
neighborhood structure among local lower bounds. Moreover, LKF17 is an efficient algorithm to calculate the HVI by partitioning the dominated space. In other words, LKF17 is also efficient in partitioning the dominated space and provides the boundary information for each hyperbox in the dominated space.

%The basic idea of the proposed algorithm to partition a higher dimensional non-dominated space is transforming the problem of partitioning a non-dominated space into the problem of partitioning the dominated space, 
The idea behind the proposed algorithm is transforming the problem of partitioning a non-dominated space into the problem of partitioning the dominated space, 
by means of introducing an intermediate Pareto-front approximation set $\pfa^{'}$. This transformation is done by the following steps. Suppose that we have a current Pareto-front approximation set $\pfa$ for a maximization problem and we want to partition the non-dominated space of $\pfa$. 
Firstly, DKLV17 is applied to locate the local lower bound points ($\mathbf{L}_d$) of $\pfa$ in the dominated space. Secondly, regard $\mathbf{L}_d$ as a new Pareto-front approximation $\pfa^{'}$ for a minimization problem with a reference point $\{\infty\}^d$. The dominated space of $\pfa^{'}$ is actually the non-dominated space of $\pfa$. Then, LKF17 can be applied to partition the dominated space of $\pfa^{'}$ by locating the lower bound points $\mathbf{l}_d$ and the upper bound points $\mathbf{u}_d$. These bound points ($\mathbf{l}_d$,$\mathbf{u}_d$) of $\pfa^{'}$ in the dominated space for a minimization problem are exact the lower/upper bound points of the partitioned, non-dominated hyperboxes of $\pfa$ for a maximization problem. The pseudo code of partitioning non-dominated space in higher dimensional cases is shown in Algorithm \ref{alg:par_nondominated_high}. 

\begin{algorithm}[ht]            
	        \KwIn{Pareto-front approximation set $\pfa$ (maximization problem), a reference point $\mathbf{r}$}
			\KwOut{Hyperboxes $S_d$}
			\nl Locate local lower bound points $\mathbf{L}_d$: $\mathbf{L}_d=DKLV17(\pfa, \mathbf{r})$\;
			\nl Set new Pareto front $\pfa^{'}$ using $\mathbf{L}_d$: $\pfa^{'}=\mathbf{L}_d$ \;
			\nl Set a new reference point $\mathbf{r}^{'}$: $\mathbf{r}^{'} = \{\infty \}^d$ \;
			\nl Get lower bound points $\mathbf{l}_d$ and upper bound points $\mathbf{u}_d$:
				$(\mathbf{l}_d,\mathbf{u}_d)=LKF17(\pfa^{'},\mathbf{r}^{'})$ \;
			\nl $S_d = (\mathbf{l}_d,\mathbf{u}_d)$ \;
			\nl Return $S_d$
			\caption{\bf Partitioning a non-dominated space in higher dimensional cases}
			\label{alg:par_nondominated_high}
\end{algorithm}

\begin{figure}[!ht]
\centering
	\includegraphics[width=0.49\textwidth]{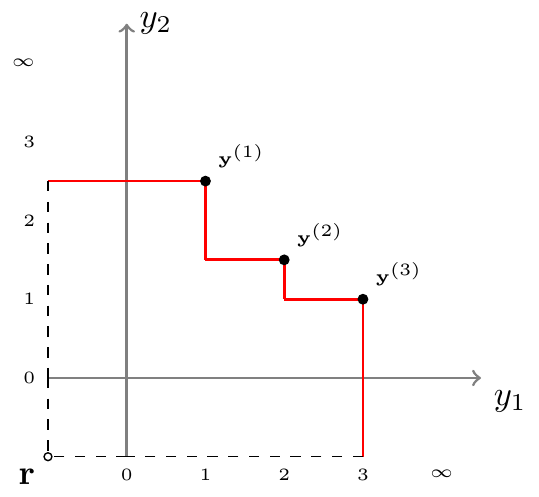}
	\includegraphics[width=0.49\textwidth]{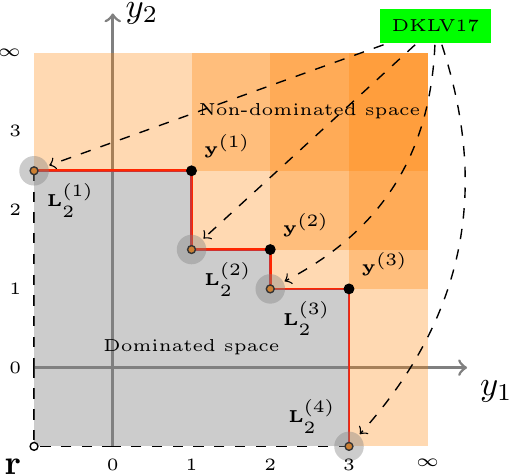}
	\includegraphics[width=0.49\textwidth]{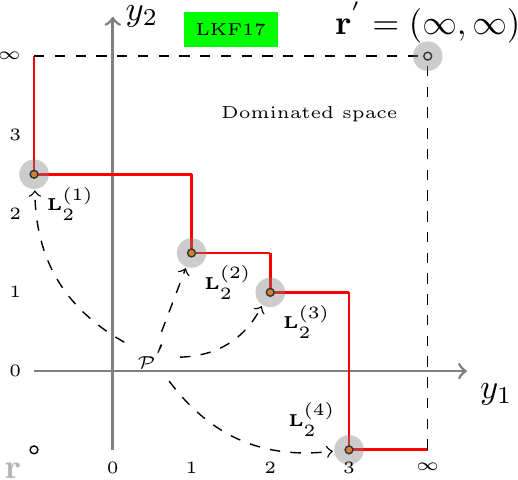}
	\includegraphics[width=0.49\textwidth]{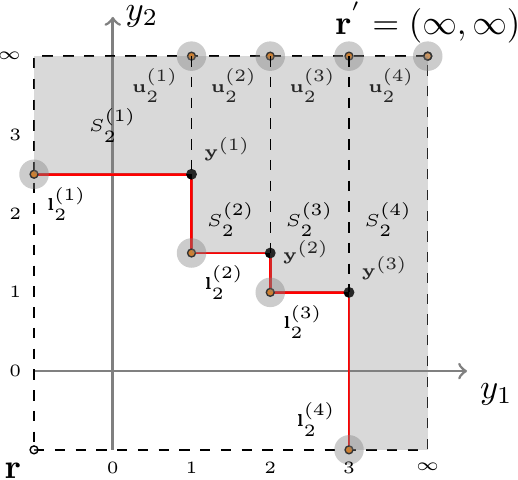}
	\caption{The illustration of partitioning a non-dominated space in higher dimensional cases. Above left: Pareto-front approximation set $\pfa$. Above right: Locating $\mathbf{L}_2$ points using DKLV17. Below left: Partitioning the dominated space of $\pfa^{'}$ using LKF17. Below right: The partitioned non-dominated space of $\pfa$.}
	\label{fig:ExamplePartHigh}
\end{figure}

\begin{example}
Figure \ref{fig:ExamplePartHigh} illustrates Algorithm \ref{alg:par_nondominated_high}. In the 2-D case, suppose a Pareto-front approximation set is $\pfa$, which consists of $\mathbf{y}^{(1)} = (1,2.5)^\top$, $\mathbf{y}^{(2)} = (2,1.5)^\top$ and $\mathbf{y}^{(3)} = (3,1)^\top$. The reference point is $\mathbf{r}=(0,0)$, see Figure \ref{fig:ExamplePartHigh} (above left). Use DKLV17 to locate the local lower bound points $\mathbf{L}_2$, which consist of $\mathbf{L}_2^{(1)} = (0,2.5)^\top$, $\mathbf{L}_2^{(2)} = (1,1.5)^\top$, $\mathbf{L}_2^{(3)} = (2,1)^\top$ and $\mathbf{L}_2^{(4)} = (3,0)^\top$, see Figure \ref{fig:ExamplePartHigh} (above right). Regard all of the local lower bound points $\mathbf{L}_2$ as the elements of a new Pareto-front approximation set $\pfa^{'}=(\mathbf{L}_2^{(1)}, \cdots, \mathbf{L}_2^{(4)})$. Set a new reference point $\mathbf{r}^{'}=(\infty,\infty)$ and utilize LKF17 to partition the dominated space of $\pfa^{'}$, by considering minimization, see Figure \ref{fig:ExamplePartHigh} (below left). The partitioned non-dominated space of $\pfa$ is then the partitioned dominated space of $\pfa^{'}$, see Figure \ref{fig:ExamplePartHigh} (below right).  
\end{example}

\subsection[title]{EHVI calculation
\footnote{Both \texttt{C++} and \texttt{MATLAB} source code for computing the EHVI are available on \url{http://liacs.leidenuniv.nl/~csmoda/index.php?page=code} or on request from the authors.}
}
\label{subsec:EHVI}
This section discusses the problem of exact EHVI calculation. Moreover, a new and efficient algorithm is also derived. Section \ref{sec:2d} and \ref{sec:3d} introduce the proposed method in the 2-D and 3-D cases, respectively. Section \ref{sec:highd} illustrates the general calculation formulas in higher dimensional cases, based on the proposed method.  

In order to simplify the notation, $\Delta(\mathbf{y})$ is used whenever $\pfa, \mathbf{r}$ are given by the context. Based on $\Delta(\mathbf{y})$, the expected hypervolume improvement function can be re-defined as:
\begin{align*}
\mbox{EHVI}(\boldsymbol\mu, \boldsymbol\sigma, \pfa, \mathbf{r}) 
&= 
\int_{\mathbb{R}^d} \mbox{HVI}(\pfa, \mathbf{y}, \mathbf{r}) \cdot \boldsymbol{\pdf}_{\boldsymbol\sigma, \boldsymbol\mu}(\mathbf{y}) d\mathbf{y}\\
&= \int_{y_1=-\infty}^\infty \cdots \int_{y_d=-\infty}^\infty \lambda_d[S_d \cap \Delta(\mathbf{y})] \boldsymbol{\pdf}_{\boldsymbol\mu, \boldsymbol\sigma}(\mathbf{y}) d\mathbf{y} \numberthis
\end{align*}\label{eq:EHVI_org}

For the convenience of expressing the EHVI formula in the remaining parts of this paper, two functions ($\ell$ and $\vartheta$) are defined as follows:
\begin{definition}[$\ell$ function]
Given the parameters of an integration slice $S_d^{(i)}$ in a $d$-dimensional space, the \emph{Hypervolume Improvement} of slice $S_d^{(i)}$ in dimension $k \leq d$ is defined as:
\begin{align*}
\ell(u_k^{(i)},y_k,l_k^{(i)}) := \lambda_1[S_d^{(i)}\cap\Delta(y_k)]=|[l_k^{(i)},u_k^{(i)}]\cap[l_k^{(i)},y_k]|=\min\{u_k^{(i)},y_k\}-l_k^{(i)}
\numberthis \label{eq:ell}
\end{align*}
\end{definition}
\begin{definition}[$\vartheta$ function]
Given the parameters of an integration slice $S_d^{(i)}$ in a $d$-dimensional space and multivariate predictive distribution $\boldsymbol\mu, \boldsymbol\sigma$, the function $\vartheta(l_k^{(i)},u_k^{(i)},\sigma_k,\mu_k)$ is then defined as: 
\begin{align*}
\vartheta(l_k^{(i)},u_k^{(i)},\sigma_k,\mu_k)
:&= \int_{y_k=u_k^{(i)}}^\infty\lambda_1[S_d^{(i)}\cap\Delta(y_k)]\cdot {\boldsymbol\pdf}_{\mu_k,\sigma_k}(y_k) dy_k \\
& =  \int_{y_k=u_k^{(i)}}^\infty (u_k^{(i)} - l_k^{(i)}) \cdot {\boldsymbol\pdf}_{\mu_k,\sigma_k}(y_k)dy_k \\
& =  (u_k^{(i)} - l_k^{(i)}) \cdot \left( 1 - \Phi \Big( \frac{u_k^{(i)}-\mu_k}{\sigma_k} \Big) \right) \qquad k=1,\cdots, d-1 \numberthis \label{eq:vartheta}
\end{align*}
\end{definition}

\subsubsection[Section title]{2-D EHVI calculation}\label{sec:2d}
According to the definition of the 2-D integration slice in Equation \ref{def:slice2d}, the \emph{Hypervolume Improvement} $\mathbf{y} \in \mathbb{R}^2$ in  the 2-D case is:
\begin{equation}
\mbox{HVI}_2(\mathbf{y},  \pfa, \mathbf{r}) = 
\sum_{i=1}^{N_2} \lambda_2 [S_2^{(i)}\cap\Delta(\mathbf{y})]
\end{equation}

$\mbox{HVI}_2$ gives rise to the compact integral for the original EHVI: 
\begin{align*}
\label{eq:limits_are_infinity}
\mbox{EHVI}(\boldsymbol{\mu},\boldsymbol{\sigma},\pfa,\mathbf{r}) = &\int_{y_1=-\infty}^\infty\int_{y_2=-\infty}^{\infty}\sum_{i=1}^{N_2} \lambda_2 [S_2^{(i)}\cap\Delta(\mathbf{y})]\cdot \boldsymbol{\pdf}_{\boldsymbol\mu,\boldsymbol\sigma}(\mathbf{y})d\mathbf{y}\numberthis
\end{align*}

Here $\mathbf{y}=(y_1,y_2)$, the intersection of $S_2^{(i)}$ with $\Delta(y_1, y_2)$ is non-empty if and only if $(\mathbf{y})$ dominates the lower left corner of $S_2^{(i)}$. 
%In other words, if and only if  $\mathbf{y}$ is located in the rectangle with lower left corner $(l_1^{(i)}, l_2^{(i)})$ and upper right corner $(\infty, \infty)$. 
Therefore:
\begin{align*}
\mbox{EHVI}(\boldsymbol{\mu},\boldsymbol{\sigma},\pfa,\mathbf{r}) = 
&\sum_{i=1}^{N_2}\int_{y_1=l_1^{(i)}}^{\infty}\int_{y_2=l_2^{(i)}}^{\infty}\lambda_2[S_2^{(i)}\cap\Delta(\mathbf{y})]\cdot \boldsymbol\pdf_{\boldsymbol\mu,\boldsymbol\sigma}(\mathbf{y})d\mathbf{y}
\label{eq:ehvi2d_initial}
\numberthis
\end{align*}

In Equation (\ref{eq:ehvi2d_initial}), the summation is done after integration. This operation is allowed, because integration is a linear mapping. Moreover, the integration interval $\int_{y_1=l_1^{(i)}}^{\infty}$ can be divided into $(\int_{y_1=l_1^{(i)}}^{u_1^{(i)}} + \int_{y_1=u_1^{(i)}}^{\infty})$, because the HVI in one dimension $\lambda_1[S_2^{(i)}\cap\Delta(y_1)]$ differs in these two integration intervals. Here $\lambda_1[B_i\cap\Delta(y_k)]$ is the HVI in dimension $k$, i.e., a 1-D HVI. Equation (\ref{eq:ehvi2d_initial}) can then be expressed as: 
\begin{align*}
\mbox{EHVI}(\boldsymbol{\mu},\boldsymbol{\sigma},\pfa,\mathbf{r}) 
&= \sum_{i=1}^{N_2}\int_{y_1=l_1^{(i)}}^{u_1^{(i)}}\int_{y_2=l_2^{(i)}}^{\infty}\lambda_2[S_2^{(i)}\cap\Delta(\mathbf{y})]\cdot \boldsymbol\pdf_{\boldsymbol\mu,\boldsymbol\sigma}(\mathbf{y})d\mathbf{y} + \numberthis \label{eq:ehvi2d_part1} \\
&\quad \sum_{i=1}^{N_2}\int_{y_1=u_1^{(i)}}^{\infty}\int_{y_2=l_2^{(i)}}^{\infty}\lambda_2[S_2^{(i)}\cap\Delta(\mathbf{y})]\cdot \boldsymbol\pdf_{\boldsymbol\mu,\boldsymbol\sigma}(\mathbf{y})d\mathbf{y} \numberthis \label{eq:ehvi2d_part2} 
\end{align*}

According to the definition of HVI, $\ell(u_1^{(i)},y_1,l_1^{(i)})$ is constant and is $(u_1^{(i)}-l_1^{(i)})$ in $\int_{y_1=u_1^{(i)}}^{\infty}$. Therefore, the \emph{Expected Improvement} in dimension $y_1$ is also a constant and it is: $\vartheta(l_1^{(i)},u_1^{(i)},\sigma_1,\mu_1)$.
Recall the $\Psi_{\infty}$ function, by which the terms (\ref{eq:ehvi2d_part1}) and (\ref{eq:ehvi2d_part2}) can be expressed as follows: 
\begin{align*}
&\text{Term }  (\ref{eq:ehvi2d_part1}) 
= 	\sum_{i=1}^{N_2}
	\left(\Psi_{\infty}(l_1^{(i)},l_1^{(i)},\mu_1,\sigma_1) - \Psi_{\infty}(l_1^{(i)},u_1^{(i)},\mu_1,\sigma_1) \right) \cdot 
	\Psi_{\infty}(l_2^{(i)},l_2^{(i)},\mu_2,\sigma_2)  \numberthis \\
&\text{Term }  (\ref{eq:ehvi2d_part2}) 
= 	\sum_{i=1}^{N_2}
	\vartheta(l_1^{(i)},u_1^{(i)},\mu_1,\sigma_1)  \cdot \Psi_{\infty}(l_2^{(i)},l_2^{(i)},\mu_2,\sigma_2) \numberthis
\end{align*}

According to Equation (\ref{eq:ehvi2d_initial}), the exact EHVI calculation needs to compute the terms (\ref{eq:ehvi2d_part1}) and (\ref{eq:ehvi2d_part2}) $n+1$ times, and each calculation requests $O(1)$ computation. To keep $\pfa$ sorted in the first coordinate requires an effort of amortized time complexity $O(\log n)$ per iteration. Hence, the time complexity of the expected hypervolume improvement in the 2-D case is in $O(n \log n)$. In the case when $\pfa$ is sorted, we can show that the time complexity is in $\Theta(n)$. 
%To do so, we will first establish a lower bound of $\Omega(n)$ for this case:
%\begin{lem}
%The computational time complexity of computing the EHVI for a set $\pfa$ that is sorted by the first coordinate is bounded from below by $\Omega(n)$.
%Proof: An adversary argument serves in \cite{Michael2016book} to prove this statement. The algorithm has to \ 'look at' all $n$ points. If a point would be omitted from the set and placed by an adversary to a new position where it changes the hypervolume indicator, this move would otherwise not be noticed by the algorithm; a move of any single point can, in general, change the expected hypervolume improvement. 
%\end{lem}

%For $d=2$, the expected improvement is computable in linear time, given that $\pfa$ is already sorted by the first coordinate.
%Next, a formula will be derived that consists of $N_2$ integrals, each of which can be solved in constant time.
\subsubsection[Section title]{3-D EHVI calculation}\label{sec:3d}
Given a partitioning of the non-dominated space into integration slices $S_3^{(1)}$, $\dots$, $S_3^{(i)}$, $\dots$, $S_3^{(2n+1)}$, 
%the part of the integral related to each of the integration slices can be computed separately.
the EHVI integrations over each slice can be computed separately. 
To see how this calculation can be done, the \emph{Hypervolume Improvement} of a point $\mathbf{y} \in \mathbb{R}^3$ is rewritten as:
\begin{equation}
\mbox{HVI}_3(\pfa, \mathbf{y},  \mathbf{r}) =
\sum_{i=1}^{N_3} \lambda_3 [S_3^{(i)}\cap\Delta(\mathbf{y})] \numberthis \label{eq:newdeltaHI}
\end{equation}
where $\Delta \mathbf{(y)}$ is the part of the objective space that is dominated by $\mathbf{y}$.
The $\mbox{HVI}$ expression in the definition of EHVI in Equation (\ref{def:EHVI}) can be replaced by $\mbox{HVI}_3$ in Equation (\ref{eq:newdeltaHI}):
\begin{align*}
\mbox{EHVI}(\boldsymbol{\mu},\boldsymbol{\sigma},\pfa,\mathbf{r}) =
\sum_{i=1}^{N_3}\int_{y_1=l_1^{(i)}}^{\infty}\int_{y_2=l_2^{(i)}}^{\infty}\int_{y_3=l_3^{(i)}}^{\infty}\lambda_3[S_3^{(i)}\cap\Delta(\mathbf{y})]\cdot \boldsymbol{\pdf}_{\boldsymbol\mu,\boldsymbol\sigma}(\mathbf{y})d\mathbf{y} \numberthis \label{eq:intlimits}
\end{align*}
Similar to the 2-D case, we can divide the integration interval $\int_{y_1=l_1^{(i)}}^{\infty}$ and $\int_{y_2=l_2^{(i)}}^{\infty}$ into $(\int_{y_1=l_1^{(i)}}^{u_1^{(i)}} + \int_{y_1=u_1^{(i)}}^{\infty})$ and $(\int_{y_2=l_2^{(i)}}^{u_2^{(i)}} + \int_{y_2=u_2^{(i)}}^{\infty})$, respectively. Also, again we can swap integration and summation based on the fact that integration is a linear mapping. Based on this subdivision, Equation (\ref{eq:intlimits}) can be expressed as:
\begin{align*}
\text{Eq.} \enskip (\ref{eq:intlimits})
&= \sum_{i=1}^{N_3}\int_{y_1=l_1^{(i)}}^{u_1^{(i)}}\int_{y_2=l_2^{(i)}}^{u_2^{(i)}}\int_{y_3=l_3^{(i)}}^{\infty}\lambda_3[S_3^{(i)}\cap\Delta(\mathbf{y})]\cdot \boldsymbol{\pdf}_{\boldsymbol\mu,\boldsymbol\sigma}(\mathbf{y})d\mathbf{y} + \numberthis \label{eq:1} \\
&\quad \sum_{i=1}^{N_3}\int_{y_1=l_1^{(i)}}^{u_1^{(i)}}\int_{y_2=u_2^{(i)}}^{\infty}\int_{y_3=l_3^{(i)}}^{\infty}\lambda_3[S_3^{(i)}\cap\Delta(\mathbf{y})]\cdot \boldsymbol{\pdf}_{\boldsymbol\mu,\boldsymbol\sigma}(\mathbf{y})d\mathbf{y} + \numberthis \label{eq:2}\\
&\quad \sum_{i=1}^{N_3}\int_{y_1=u_1^{(i)}}^{\infty}\int_{y_2=l_2^{(i)}}^{u_2^{(i)}}\int_{y_3=l_3^{(i)}}^{\infty}\lambda_3[S_3^{(i)}\cap\Delta(\mathbf{y})]\cdot \boldsymbol{\pdf}_{\boldsymbol\mu,\boldsymbol\sigma}(\mathbf{y})d\mathbf{y} + \numberthis \label{eq:3}\\
&\quad \sum_{i=1}^{N_3}\int_{y_1=u_1^{(i)}}^{\infty}\int_{y_2=u_2^{(i)}}^{\infty}\int_{y_3=l_3^{(i)}}^{\infty}\lambda_3[S_3^{(i)}\cap
\Delta(\mathbf{y})]\cdot \boldsymbol{\pdf}_{\boldsymbol\mu,\boldsymbol\sigma}(\mathbf{y})d\mathbf{y} \numberthis    \label{eq:4}
\end{align*}
Recalling the definition of the $\vartheta$ function and calculation of $\lambda_1[B_i\cap\Delta(y_k)]$, the term (\ref{eq:1}) can be rewritten as follows:

\begin{align*}
\text{Term} \enskip (\ref{eq:1})
&=\sum_{i=1}^{N_3}
(\Psi_{\infty}(l_1^{(i)},l_1^{(i)},\mu_1,\sigma_1) - \Psi_{\infty}(l_1^{(i)},u_1^{(i)},\sigma_1,\mu_1))\cdot \\
&\qquad \quad
(\Psi_{\infty}(l_2^{(i)},l_2^{(i)},\mu_2,\sigma_2) - \Psi_{\infty}(l_2^{(i)},u_2^{(i)},\sigma_2,\mu_2)) \cdot
\Psi_{\infty}(l_3^{(i)},l_3^{(i)},\mu_3,\sigma_3) \numberthis \label{eq:1_new}
\end{align*}
Similar to the derivation of the term (\ref{eq:1}), the terms (\ref{eq:2}), (\ref{eq:3}) and (\ref{eq:4}) can be written as follows:
\begin{align*}
\text{Term} \enskip (\ref{eq:2})
&= \sum_{i=1}^{N_3} ( \Psi_{\infty}(l_1^{(i)},l_1^{(i)},\mu_1,\sigma_1) - \Psi_{\infty}(l_1^{(i)},u_1^{(i)},\sigma_1,\mu_1)) \cdot \vartheta(l_2^{(i)},u_2^{(i)},\sigma_2,\mu_2) \cdot \\
&\qquad \quad \Psi_{\infty}(l_3^{(i)},l_3^{(i)},\mu_3,\sigma_3)  \numberthis \label{eq:2_new} \\
\text{Term} \enskip (\ref{eq:3})
&= \sum_{i=1}^{N_3} \vartheta(l_1^{(i)},u_1^{(i)},\sigma_1,\mu_1) \cdot
( \Psi_{\infty}(l_2^{(i)},l_2^{(i)},\mu_2,\sigma_2) - \Psi_{\infty}(l_2^{(i)},u_2^{(i)},\sigma_2,\mu_2))  \cdot \\
&\qquad \quad \Psi_{\infty}(l_3^{(i)},l_3^{(i)},\mu_3,\sigma_3) \numberthis \label{eq:3_new} \\
\text{Term} \enskip (\ref{eq:4})
&= \sum_{i=1}^{N_3} \vartheta(l_1^{(i)},u_1^{(i)},\sigma_1,\mu_1) \cdot \vartheta(l_2^{(i)},u_2^{(i)},\sigma_2,\mu_2) \cdot \Psi_{\infty}(l_3^{(i)},l_3^{(i)},\mu_3,\sigma_3) \numberthis \label{eq:4_new}
\end{align*}
The final EHVI formula is the sum of the terms (\ref{eq:1_new}), (\ref{eq:2_new}), (\ref{eq:3_new}) and (\ref{eq:4_new}). 

%During the EHVI calculation, as the $y_1y_2$-projections are mutually non-dominated, the points are also sorted by the $y_2$ coordinate in the AVL tree, identifying a neighboring point or a discard point takes time $O(\log n)$. 
During the EHVI calculation, $y_1y_2$-projections are mutually non-dominated. Moreover, the points are sorted by the $y_2$ coordinate in the AVL tree. Therefore,  identifying a neighboring/discard point takes time $O(\log n)$. Then the EHVI for these integration slices is calculated by the summation of the term (\ref{eq:1_new}), (\ref{eq:2_new}), (\ref{eq:3_new}) and (\ref{eq:4_new}), with the parameters of $\boldsymbol{\mu}, \boldsymbol{\sigma}$ and $S_3^{(N_3)}$. The EHVI time complexity for each slice is $O(1)$. Moreover, the dominated points ($\mathbf{y}^{\text{(d[s])}}$) are removed from the AVL tree, and the new points ($\mathbf{y}^{(j)}$) are inserted into the AVL tree. Since the points dominated by the new point $\mathbf{y}^{(j)}$ are deleted at the end of the current loop, they will not occur again in later computations. Hence, the total number of open slices does not exceed $N_3$, as mentioned before, and the total computational cost is $O(n \log n)$. 
\subsubsection[Section title]{Higher dimensional EHVI}\label{sec:highd}
The interval of integration in each coordinate (except the last) can be divided into 2 parts: $[l,u]$ and $[u,\infty]$. Therefore, the equation for EHVI for each hyperbox can be decomposed into $2^{d-1}$ parts. For the interval of $[u,\infty]$, the improvements ($\lambda_k[S_d^{(i)}\cap\Delta(y_k)]$) are constant values, and the $\Psi_{\infty}$ function can be simplified by calculating function $\Phi$ and the improvement in these coordinates. For the last coordinate, there is no need to separate the interval, because the improvement in this coordinate ($\lambda_d[S_d^{(i)}\cap\Delta(y_d)]$) is a variable in $[l,\infty]$.

According to the definition of higher dimensional integral boxes in Section \ref{subsec:ehvi_hd}, the EHVI ($d\geqslant4$) can be calculated by the following equation:

\begin{landscape}
\begin{align*}
&\mbox{EHVI}(\boldsymbol{\mu},\boldsymbol{\sigma},\pfa,\mathbf{r})\\ 
&= \sum_{i=1}^{N_d}\int_{y_1=l_1^{(i)}}^{\infty} \cdots \int_{y_d=l_d^{(i)}}^{\infty} \lambda_d[S_d^{(i)}\cap\Delta(y_1,\cdots,y_d)]\cdot \boldsymbol{\pdf}_{\boldsymbol\mu,\boldsymbol\sigma}(\mathbf{y})d\mathbf{y} \\
&=\sum_{i=1}^{N_d}\left( (\int_{y_1=l_1^{(i)}}^{y_1=u_1^{(i)}} + \int_{y_1=u_1^{(i)}}^{\infty}) \cdots (\int_{y_{d-1}=l_{d-1}^{(i)}}^{y_{d-1}=u_{d-1}^{(i)}} + \int_{y_{d-1}=u_{d-1}^{(i)}}^{\infty}) \cdot \int_{y_d=l_d^{(i)}}^{\infty} \right) \cdot \lambda_d[S_d^{(i)}\cap\Delta(y_1,\cdots,y_d)]\cdot \boldsymbol{\pdf}_{\boldsymbol\mu,\boldsymbol\sigma}(\mathbf{y})d\mathbf{y}  \\
%----------------------------------------------------
&=
\begin{pmatrix}
  \int_{y_1=l_1^{(i)}}^{u_1^{(i)}} & \int_{y_2=l_2^{(i)}}^{u_2^{(i)}} & \cdots  
  & \int_{y_{d-2}=l_{d-2}^{(i)}}^{u_{d-2}^{(i)}} & \int_{y_{d-1}=l_{d-1}^{(i)}}^{u_{d-1}^{(i)}} & \int_{y_d=l_d^{(i)}}^{u_d^{(i)}} \\
  \int_{y_1=l_1^{(i)}}^{u_1^{(i)}} & \int_{y_2=l_2^{(i)}}^{u_2^{(i)}} & \cdots  
  & \int_{y_{d-2}=l_{d-2}^{(i)}}^{u_{d-2}^{(i)}} & \int_{y_{d-1}=l_{d-1}^{(i)}}^{\infty} & \int_{y_d=l_d^{(i)}}^{u_d^{(i)}} \\
  \int_{y_1=l_1^{(i)}}^{u_1^{(i)}} & \int_{y_2=l_2^{(i)}}^{u_2^{(i)}} & \cdots  
  & \int_{y_{d-2}=l_{d-2}^{(i)}}^{\infty} & \int_{y_{d-1}=l_{d-1}^{(i)}}^{u_{d-1}^{(i)}} & \int_{y_d=l_d^{(i)}}^{u_d^{(i)}} \\
  \int_{y_1=l_1^{(i)}}^{u_1^{(i)}} & \int_{y_2=l_2^{(i)}}^{u_2^{(i)}} & \cdots  
  & \int_{y_{d-2}=l_{d-2}^{(i)}}^{\infty} & \int_{y_{d-1}=l_{d-1}^{(i)}}^{\infty} & \int_{y_d=l_d^{(i)}}^{u_d^{(i)}} \\
  \vdots  & \vdots  & \ddots & \vdots  & \vdots & \vdots \\
  \int_{y_1=l_1^{(i)}}^{\infty} & \int_{y_2=l_2^{(i)}}^{\infty} & \cdots  
  & \int_{y_{d-2}=l_{d-2}^{(i)}}^{u_{d-2}^{(i)}} & \int_{y_{d-1}=l_{d-1}^{(i)}}^{\infty} & \int_{y_d=l_d^{(i)}}^{u_d^{(i)}} \\
  \int_{y_1=l_1^{(i)}}^{\infty} & \int_{y_2=l_2^{(i)}}^{\infty} & \cdots  
  & \int_{y_{d-2}=l_{d-2}^{(i)}}^{u_{d-2}^{(i)}} & \int_{y_{d-1}=l_{d-1}^{(i)}}^{u_{d-1}^{(i)}} & \int_{y_d=l_d^{(i)}}^{u_d^{(i)}} \\
  \int_{y_1=l_1^{(i)}}^{\infty} & \int_{y_2=l_2^{(i)}}^{\infty} & \cdots  
  & \int_{y_{d-2}=l_{d-2}^{(i)}}^{\infty} & \int_{y_{d-1}=l_{d-1}^{(i)}}^{u_{d-1}^{(i)}} & \int_{y_d=l_d^{(i)}}^{u_d^{(i)}} \\
  \int_{y_1=l_1^{(i)}}^{\infty} & \int_{y_2=l_2^{(i)}}^{\infty} & \cdots  
  & \int_{y_{d-2}=l_{d-2}^{(i)}}^{\infty} & \int_{y_{d-1}=l_{d-1}^{(i)}}^{\infty} & \int_{y_d=l_d^{(i)}}^{u_d^{(i)}} \\
\end{pmatrix} 
\lambda_d[S_d^{(i)}\cap\Delta(y_1,\cdots,y_d)]\cdot \boldsymbol{\pdf}_{\boldsymbol\mu,\boldsymbol\sigma}(\mathbf{y})d\mathbf{y} \\
\end{align*}
\begin{align*}
&=
\begin{pmatrix}
  \binom{\Psi_\infty(l_{1}^{(i)},l_{1}^{(i)},\mu_{1},\sigma_{1}) -}{\Psi_\infty(l_{1}^{(i)},u_{1}^{(i)},\sigma_{1},\mu_{1})}
& \binom{\Psi_\infty(l_{2}^{(i)},l_{2}^{(i)},\mu_{2},\sigma_{2}) -}{\Psi_\infty(l_{2}^{(i)},u_{2}^{(i)},\sigma_{2},\mu_{2})}
& \cdots 
& \binom{\Psi_\infty(l_{d-2}^{(i)},l_{d-2}^{(i)},\mu_{d-2},\sigma_{d-2}) -}{\Psi_\infty(l_{d-2}^{(i)},u_{d-2}^{(i)},\sigma_{d-2},\mu_{d-2})}
& \binom{\Psi_\infty(l_{d-1}^{(i)},l_{d-1}^{(i)},\mu_{d-1},\sigma_{d-1}) -}{\Psi_\infty(l_{d-1}^{(i)},u_{d-1}^{(i)},\sigma_{d-1},\mu_{d-1})}
& \Psi_{\infty}(l_d^{(i)},l_d^{(i)},\mu_d,\sigma_d)  
& + \\
  \binom{\Psi_\infty(l_{1}^{(i)},l_{1}^{(i)},\mu_{1},\sigma_{1}) -}{\Psi_\infty(l_{1}^{(i)},u_{1}^{(i)},\sigma_{1},\mu_{1})}
& \binom{\Psi_\infty(l_{2}^{(i)},l_{2}^{(i)},\mu_{2},\sigma_{2}) -}{\Psi_\infty(l_{2}^{(i)},u_{2}^{(i)},\sigma_{2},\mu_{2})}
& \cdots 
& \binom{\Psi_\infty(l_{d-2}^{(i)},l_{d-2}^{(i)},\mu_{d-2},\sigma_{d-2}) -}{\Psi_\infty(l_{d-2}^{(i)},u_{d-2}^{(i)},\sigma_{d-2},\mu_{d-2})}
& \vartheta(l_{d-1}^{(i)},u_{d-1}^{(i)},\sigma_{d-1},\mu_{d-1}) 
& \Psi_{\infty}(l_d^{(i)},l_d^{(i)},\mu_d,\sigma_d)  
& + \\
  \binom{\Psi_\infty(l_{1}^{(i)},l_{1}^{(i)},\mu_{1},\sigma_{1}) -}{\Psi_\infty(l_{1}^{(i)},u_{1}^{(i)},\sigma_{1},\mu_{1})}
& \binom{\Psi_\infty(l_{2}^{(i)},l_{2}^{(i)},\mu_{2},\sigma_{2}) -}{\Psi_\infty(l_{2}^{(i)},u_{2}^{(i)},\sigma_{2},\mu_{2})}
& \cdots 
& \vartheta(l_{d-2}^{(i)},u_{d-2}^{(i)},\sigma_{d-2},\mu_{d-2})
& \binom{\Psi_\infty(l_{d-1}^{(i)},l_{d-1}^{(i)},\mu_{d-1},\sigma_{d-1}) -}{\Psi_\infty(l_{d-1}^{(i)},u_{d-1}^{(i)},\sigma_{d-1},\mu_{d-1})}
& \Psi_{\infty}(l_d^{(i)},l_d^{(i)},\mu_d,\sigma_d)  
& + \\
  \binom{\Psi_\infty(l_{1}^{(i)},l_{1}^{(i)},\mu_{1},\sigma_{1}) -}{\Psi_\infty(l_{1}^{(i)},u_{1}^{(i)},\sigma_{1},\mu_{1})}
& \binom{\Psi_\infty(l_{2}^{(i)},l_{2}^{(i)},\mu_{2},\sigma_{2}) -}{\Psi_\infty(l_{2}^{(i)},u_{2}^{(i)},\sigma_{2},\mu_{2})}
& \cdots 
& \vartheta(l_{d-2}^{(i)},u_{d-2}^{(i)},\sigma_{d-2},\mu_{d-2})
& \vartheta(l_{d-1}^{(i)},u_{d-1}^{(i)},\sigma_{d-1},\mu_{d-1}) 
& \Psi_{\infty}(l_d^{(i)},l_d^{(i)},\mu_d,\sigma_d)  \\
%-----------------------------------------------------------------
\vdots  & \vdots  & \ddots & \vdots  & \vdots & \vdots & \vdots \\
  \vartheta(l_{1}^{(i)},u_{1}^{(i)},\sigma_{1},\mu_{1}) 
& \vartheta(l_{2}^{(i)},u_{2}^{(i)},\sigma_{2},\mu_{2})
& \cdots 
& \binom{\Psi_\infty(l_{d-2}^{(i)},l_{d-2}^{(i)},\mu_{d-2},\sigma_{d-2}) -}{\Psi_\infty(l_{d-2}^{(i)},u_{d-2}^{(i)},\sigma_{d-2},\mu_{d-2})}
& \binom{\Psi_\infty(l_{d-1}^{(i)},l_{d-1}^{(i)},\mu_{d-1},\sigma_{d-1}) -}{\Psi_\infty(l_{d-1}^{(i)},u_{d-1}^{(i)},\sigma_{d-1},\mu_{d-1})}
& \Psi_{\infty}(l_d^{(i)},l_d^{(i)},\mu_d,\sigma_d)  
& + \\
  \vartheta(l_{1}^{(i)},u_{1}^{(i)},\sigma_{1},\mu_{1}) 
& \vartheta(l_{2}^{(i)},u_{2}^{(i)},\sigma_{2},\mu_{2})
& \cdots 
& \binom{\Psi_\infty(l_{d-2}^{(i)},l_{d-2}^{(i)},\mu_{d-2},\sigma_{d-2}) -}{\Psi_\infty(l_{d-2}^{(i)},u_{d-2}^{(i)},\sigma_{d-2},\mu_{d-2})}
& \vartheta(l_{d-1}^{(i)},u_{d-1}^{(i)},\sigma_{d-1},\mu_{d-1}) 
& \Psi_{\infty}(l_d^{(i)},l_d^{(i)},\mu_d,\sigma_d)  
& + \\
  \vartheta(l_{1}^{(i)},u_{1}^{(i)},\sigma_{1},\mu_{1}) 
& \vartheta(l_{2}^{(i)},u_{2}^{(i)},\sigma_{2},\mu_{2})
& \cdots 
& \vartheta(l_{d-2}^{(i)},u_{d-2}^{(i)},\sigma_{d-2},\mu_{d-2})
& \binom{\Psi_\infty(l_{d-1}^{(i)},l_{d-1}^{(i)},\mu_{d-1},\sigma_{d-1}) -}{\Psi_\infty(l_{d-1}^{(i)},u_{d-1}^{(i)},\sigma_{d-1},\mu_{d-1})}
& \Psi_{\infty}(l_d^{(i)},l_d^{(i)},\mu_d,\sigma_d)  
& + \\
  \vartheta(l_{1}^{(i)},u_{1}^{(i)},\sigma_{1},\mu_{1}) 
& \vartheta(l_{2}^{(i)},u_{2}^{(i)},\sigma_{2},\mu_{2})
& \cdots 
& \vartheta(l_{d-2}^{(i)},u_{d-2}^{(i)},\sigma_{d-2},\mu_{d-2})
& \vartheta(l_{d-1}^{(i)},u_{d-1}^{(i)},\sigma_{d-1},\mu_{d-1}) 
& \Psi_{\infty}(l_d^{(i)},l_d^{(i)},\mu_d,\sigma_d)  
\end{pmatrix} \label{eq:ehvi_high_original} \numberthis\\
& =
\begin{pmatrix}
\omega(i,1,0) & \omega(i,2,0) & \cdots & \omega(i,d-2,0) & \omega(i,d-1,0) & \Psi_{\infty}(l_d^{(i)},l_d^{(i)},\mu_d,\sigma_d) & + \\
\omega(i,1,0) & \omega(i,2,0) & \cdots & \omega(i,d-2,0) & \omega(i,d-1,1) & \Psi_{\infty}(l_d^{(i)},l_d^{(i)},\mu_d,\sigma_d) & + \\
\omega(i,1,0) & \omega(i,2,0) & \cdots & \omega(i,d-2,1) & \omega(i,d-1,0) & \Psi_{\infty}(l_d^{(i)},l_d^{(i)},\mu_d,\sigma_d) & + \\
\omega(i,1,0) & \omega(i,2,0) & \cdots & \omega(i,d-2,1) & \omega(i,d-1,1) & \Psi_{\infty}(l_d^{(i)},l_d^{(i)},\mu_d,\sigma_d) & + \\
%-----------------------------------------------------------------
\vdots  & \vdots  & \ddots & \vdots  & \vdots & \vdots & \vdots \\
\omega(i,1,1) & \omega(i,2,1) & \cdots & \omega(i,d-2,0) & \omega(i,d-1,0) & \Psi_{\infty}(l_d^{(i)},l_d^{(i)},\mu_d,\sigma_d)  & + \\
\omega(i,1,1) & \omega(i,2,1) & \cdots & \omega(i,d-2,0) & \omega(i,d-1,1) & \Psi_{\infty}(l_d^{(i)},l_d^{(i)},\mu_d,\sigma_d)  & + \\
\omega(i,1,1) & \omega(i,2,1) & \cdots & \omega(i,d-2,1) & \omega(i,d-1,0) & \Psi_{\infty}(l_d^{(i)},l_d^{(i)},\mu_d,\sigma_d)  & + \\
\omega(i,1,1) & \omega(i,2,1) & \cdots & \omega(i,d-2,1) & \omega(i,d-1,1) & \Psi_{\infty}(l_d^{(i)},l_d^{(i)},\mu_d,\sigma_d)  \\
\end{pmatrix}  \\
\end{align*}
\begin{align*}
&=
\begin{pmatrix}
\omega(i,1,C^{(0)_2}_1) & \omega(i,2,C^{(0)_2}_2) & \cdots & \omega(i,d-2,C^{(0)_{2}}_{(d-2)}) & \omega(i,d-1,C^{(0)_{2}}_{(d-1)}) & \Psi_{\infty}(l_d^{(i)},l_d^{(i)},\mu_d,\sigma_d) & + \\
\omega(i,1,C^{(1)_2}_1) & \omega(i,2,C^{(1)_2}_2) & \cdots & \omega(i,d-2,C^{(1)_{2}}_{(d-2)}) & \omega(i,d-1,C^{(1)_{2}}_{(d-1)}) & \Psi_{\infty}(l_d^{(i)},l_d^{(i)},\mu_d,\sigma_d) & + \\
\omega(i,1,C^{(2)_2}_1) & \omega(i,2,C^{(2)_2}_2) & \cdots & \omega(i,d-2,C^{(2)_{2}}_{(d-2)}) & \omega(i,d-1,C^{(2)_{2}}_{(d-1)}) & \Psi_{\infty}(l_d^{(i)},l_d^{(i)},\mu_d,\sigma_d) & + \\
\omega(i,1,C^{(3)_2}_1) & \omega(i,2,C^{(3)_2}_2) & \cdots & \omega(i,d-2,C^{(3)_{2}}_{(d-2)}) & \omega(i,d-1,C^{(3)_{2}}_{(d-1)}) & \Psi_{\infty}(l_d^{(i)},l_d^{(i)},\mu_d,\sigma_d) & + \\
%-----------------------------------------------------------------
\vdots  & \vdots  & \ddots & \vdots  & \vdots & \vdots & \vdots \\
\omega(i,1,C^{(2^{d-1}-4)_2}_1) & \omega(i,2,C^{(2^{d-1}-4)_2}_2) & \cdots & \omega(i,d-2,C^{(2^{d-1}-4)_{2}}_{d-2}) & \omega(i,d-1,C^{(2^{d-1}-4)_{2}}_{d-1}) & \Psi_{\infty}(l_d^{(i)},l_d^{(i)},\mu_d,\sigma_d)  & + \\
\omega(i,1,C^{(2^{d-1}-3)_2}_1) & \omega(i,2,C^{(2^{d-1}-3)_2}_2) & \cdots & \omega(i,d-2,C^{(2^{d-1}-3)_{2}}_{d-2}) & \omega(i,d-1,C^{(2^{d-1}-3)_{2}}_{d-1}) & \Psi_{\infty}(l_d^{(i)},l_d^{(i)},\mu_d,\sigma_d)  & + \\
\omega(i,1,C^{(2^{d-1}-2)_2}_1) & \omega(i,2,C^{(2^{d-1}-2)_2}_2) & \cdots & \omega(i,d-2,C^{(2^{d-1}-2)_{2}}_{d-2}) & \omega(i,d-1,C^{(2^{d-1}-2)_{2}}_{d-1}) & \Psi_{\infty}(l_d^{(i)},l_d^{(i)},\mu_d,\sigma_d)  & + \\
\omega(i,1,C^{(2^{d-1}-1)_2}_1) & \omega(i,2,C^{(2^{d-1}-1)_2}_2) & \cdots & \omega(i,d-2,C^{(2^{d-1}-1)_{2}}_{d-2}) & \omega(i,d-1,C^{(2^{d-1}-1)_{2}}_{d-1}) & \Psi_{\infty}(l_d^{(i)},l_d^{(i)},\mu_d,\sigma_d)  \\
\end{pmatrix}  \\
&= \sum_{i=1}^{N_d} 
\overbrace{
\left( 
\sum_{j=0}^{2^{d-1}-1} 
	\left( \underbrace{
	\prod_{k=1}^{d-1} \omega(i,k,C_k^{(j)_2}) \cdot \Psi_{\infty}(l_d^{(i)},l_d^{(i)},\mu_d,\sigma_d)}_{\text{An integration for a divided interval.}}
	\right)  
\right) 
}^{\text{The EHVI value for a hyperbox $S_d^{(i)}$, denoted as $EHVI_{S_d^{(i)}}$}.}
\label{eq:high_ehvi} \numberthis
\end{align*}
\end{landscape}

In the term (\ref{eq:ehvi_high_original}), the integral of each dimension $\int_{y_k=l_k^{(i)}}^{u_k^{(i)}}\lambda_k[S_k^{(i)}\cap\Delta(y_1,\cdots,y_k)]\cdot \boldsymbol\pdf$ 
$_{\boldsymbol\mu,\boldsymbol\sigma}(\mathbf{y})d\mathbf{y}|_{1 \leq k \leq d-1}$ has two and only two different expressions ($\Psi_\infty$ or $\vartheta$), except that in the last dimension ($k=d$), the expression of $\int_{y_d=l_d^{(i)}}^{u_d^{(i)}}\lambda_d[S_d^{(i)}\cap\Delta(y_1,\cdots,y_d)]\cdot \boldsymbol{\pdf}_{\boldsymbol\mu,\boldsymbol\sigma}(\mathbf{y})d\mathbf{y}|_{k=d}$ is always $\Psi_\infty$. 
The final expression of the EHVI is the sum of the $EHVI_{S_d^{(i)}}$ in Equation (\ref{eq:high_ehvi}) for all the partitioned, non-dominated hyperboxes $S_d^{(i)}|_{i=1,\cdots,N_d}$. In Equation (\ref{eq:high_ehvi}), $EHVI_{S_d^{(i)}}$ has $2^{d-1}$ terms because the integration has two different expressions in dimension $k$ (${1 \leq k \leq d-1}$) and has only one expression in $d$-th dimension ($k=d$).  
%The final expression of EHVI is the sum of the combination of the product of the different expressions. Since the integral of dimension $1 \leq k \leq d-1$ has two different expressions and dimension $k=d$ has one expression, the final EHVI expression is the sum of $2^{d-1}$ terms. 

In Equation (\ref{eq:high_ehvi}), $(j)_2$ stands for the binary string representation of the integer $j$. The length of $(j)_2$ is $d-1$. $C^{(j)_2}_k$ is a bit and represents the $k$-th bit of $(j)_2$ in the binary string. For example, if $d=5$, $j=8$, then $(j)_2=(1 \enskip 0 \enskip 0 \enskip 0)$, $C^{(j)_2}_{k=4} = 1$ and $C^{(j)_2}_{k=1,2,3} = 0$. In Equation (\ref{eq:high_ehvi}), $\omega(i,k,C_k^{(j)_2})$ is defined as:\\
\begin{align*}
&\omega(i,k,C_k^{(j)_2}) := 
\begin{cases}
               \Psi_\infty(l_k^{(i)},l_k^{(i)},\mu_k,\sigma_k) - \Psi_\infty(l_k^{(i)},u_k^{(i)},\sigma_k,\mu_k) 
               			\enskip \mbox{if} \enskip C_k^{(j)_2}=0 \\      
               \vartheta(l_k^{(i)},u_k^{(i)},\sigma_k,\mu_k)   
               			\hspace{3.82cm} \mbox{if} \enskip C_k^{(j)_2}=1 \\
\end{cases} \numberthis
\label{eq:high_ehvi_final}
\end{align*}

Equation (\ref{eq:high_ehvi}) shows how to calculate EHVI in the case of $d$ objectives. According to Equation (\ref{eq:high_ehvi}), the runtime complexity of the proposed algorithm can be calculated. The exact EHVI is given by $\sum_{j=0}^{2^{d-1}-1} \left( \prod_{k=1}^{d-1} \omega(i,k,C_k^{(j)_2}) \cdot \Psi_{\infty}(l_d^{(i)},l_d^{(i)},\mu_d,\sigma_d) \right)$, which requires $O(1)$ computation steps for each hyperbox calculation. Currently, the exact number of hyperboxes $N_d|_{d\geq4}$ for a non-dominated space is still unknown. 
It is hypothesized by the authors that $N_d$ is the exact number of the local lower bound points, which can be calculated by the DKLV17 algorithm. The LKF17 algorithm partitions the non-dominated space into $\Theta(n^{\lfloor d/2\rfloor})$ hyperboxes and the time complexity for computing these hyperboxes grows linearly with the number of boxes (see, e.g., Lacour et al. \cite{lacour2017box}). Given a fixed dimension $d$, the time complexity of our EHVI computation algorithm is therefore also in $\Theta(n^{\lfloor d/2\rfloor})$. 

Note that every hyperbox requires $O(2^{d-1})$ time complexity. Therefore, the complexity in terms of $d$ and $n$ is given by $O(2^{d-1} \cdot n^{\lfloor d/2\rfloor})$. Due to the exponential dependence on $d$, the EHVI computation algorithm is only useful in moderate dimensional cases. Note, however, that a much faster computation cannot be expected, as the time complexity of the hypervolume indicator itself scales superpolynomially with the number of objectives $d$, under the assumption P $\neq$ NP. It is easy to show, that the EHVI computation has at least the same time complexity than the hypervolume indicator computation \cite{Michael2016book}, and it is therefore also an NP hard problem in $d$ -- but polynomial in $n$ for any fixed value of $d$.
%The upper bound of runtime complexity is $O(n\tau)$, where $O(\tau)$ is the computation complexity of the search algorithm. For the case of $d=1,2,3$, $O(\tau) \in O(\log n)$. 

%CH -- below, change "can not finish" to "cannot finish"

\subsection{Probability of Improvement (PoI)}
According to the partitioning method in Section \ref{subsec:Partitioning}, $\mbox{PoI}$ can be calculated as follows: 
\begin{align*}
\label{eq:poiparts}
\mbox{PoI}(\boldsymbol{\mu},\boldsymbol{\sigma},\pfa) 
& = \int_{y_1=-\infty}^\infty \cdots \int_{y_d=-\infty}^\infty \mathrm{I}(\mathbf{y} \mbox{ impr } \pfa ) \boldsymbol{\pdf}_{\boldsymbol\mu, \boldsymbol\sigma}(\mathbf{y}) \mathrm{d}y_1 \dots \mathrm{d}y_d \\
& =  \sum_{i=1}^{N_d} \prod_{j=1}^{d} \mathrm{I}(\mathbf{y} \mbox{ impr } \pfa ) \Big( \Phi (\frac{u_j^{(i)}-\mu_j}{\sigma_j}) - \Phi(\frac{l_j^{(i)}-\mu_j}{\sigma_j}) \Big) \numberthis
\end{align*}
Here, $N_d$ is the number of integration slices, and $N_2=n+1$, $N_3=2n+1$ in the 2-D and 3-D case respectively. Since $\mbox{PoI}$ is a reference-free indicator, 
%\footnote{This means the integral space for $\mbox{PoI}$ is the whole non-dominated space, which dominates $\pfa$.}, 
a reference point $\mathbf{r}=\{ - \infty \}^d$ is only used in order to obtain the correct boundary information ($\mathbf{l}_d,\mathbf{u}_d$).  
\section{Experiments}\label{sec:exp}
\subsection{Speed Comparison}
The test benchmarks from Emmerich and Fonseca \cite{emmerich2011computing} were used to generate Pareto-front sets. The Pareto-front sets and evaluated points were randomly generated based on {\sc convexSpherical} and {\sc concaveSpherical} functions. Two EHVI calculation algorithms, CDD13 \cite{couckuyt2014fast} and KMAC, were compared using the same benchmarks in this experiment.\footnote{Another EHVI calculation algorithm, IRS\_fast, is not compared in this paper, as it only works when $d=2,3$. For the detailed speed comparisons between IRS\_fast and KMAC in the 3-D case, see \cite{yang2017computing}.} Note that KMAC algorithm in this paper includes the KMAC\_2D for 2-D EHVI calculation in \cite{Michael2016book}, the KMAC\_3D algorithm for 3-D EHVI calculation in \cite{yang2017computing}, and the extended KMAC algorithm for higher dimensional cases ($d\geq4$) in this paper. 

The parameters: $\sigma_d = 2.5$, $\mu_d = 10, d = 2, \cdots, 5$ are used in the experiments. Pareto front sizes are $|\pfa|\in{\lbrace10, 20, \cdots, 200 \rbrace}$ and Batch Size is 1, which represents the number of the evaluated points under the same Pareto-front approximation set. Ten $\pfa$ sets are randomly generated by the same parameters. Average runtimes over 100 repetitions (10 repetitions for 10 $\pfa$ sets) were computed. All the experiments were performed on the same computer, and the hardware is: Intel(R) Xeon(R) CPU I7 3770 3.40GHz, RAM 16GB. The operating system is Ubuntu 16.04 LTS (64 bit), the compiler of KMAC is g++ 4.9.2 with compiler flag -Ofast, and CDD13 is based on MATLAB 8.4.0.150421 (R2014b), 64 bit. The experiments are set to halt when the algorithms cannot finish the EHVI computation within 30 minutes. 

\begin{figure}[!ht]
    \begin{subfigure}[ht]{1\textwidth}
        \centering
        \includegraphics[width=1\textwidth]{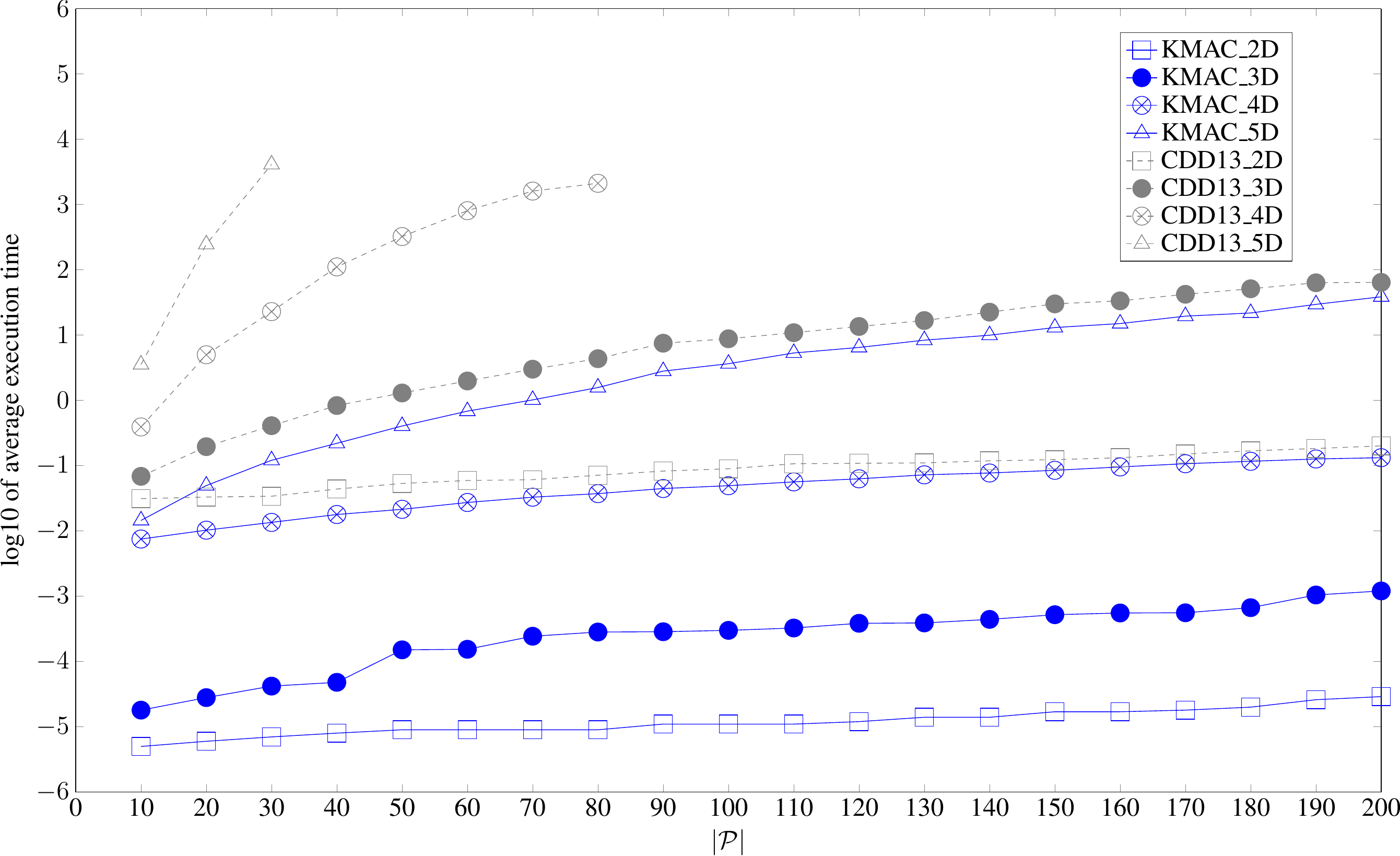}
    \end{subfigure}
    \begin{subfigure}[ht]{1\textwidth}
    	\centering
		\includegraphics[width=1\textwidth]{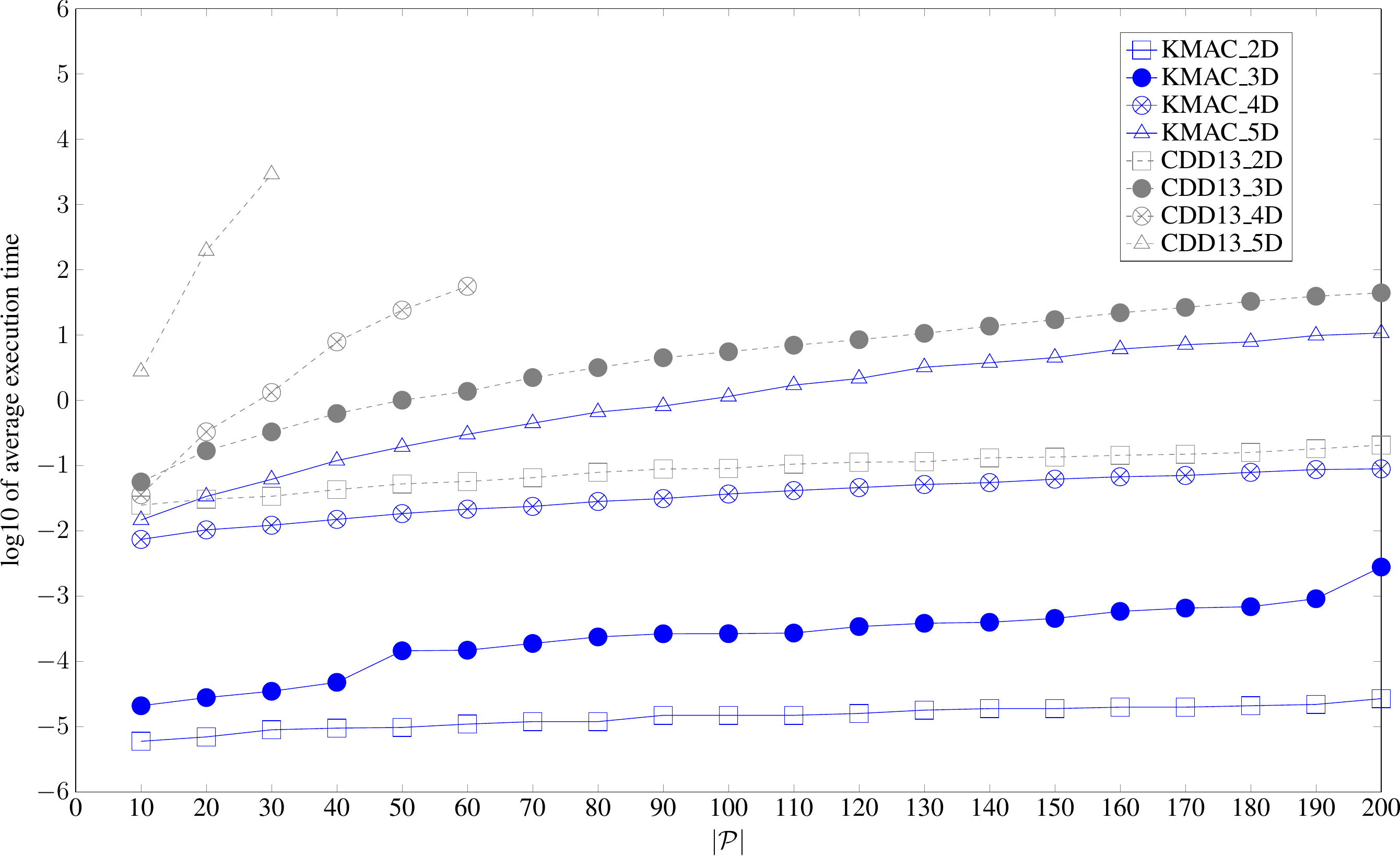}
    \end{subfigure}
    \caption{\label{fig:exp_high23}Speed comparison of EHVI calculation. Above: concave random Pareto-front set; Below: convex random Pareto-front set.}
\end{figure}

%CH -- below, change "all these algorithms" to "all of these algorithms"

The experimental results in Figure \ref{fig:exp_high23} show that KMAC is much faster than CDD13, especially when $|\pfa|$ is increased. Moreover, CDD13 can not calculate the exact EHVI value within 30 minutes when $|\pfa|$ is bigger than 30 in the 5-D case. 
\subsection{Benchmark Performance}\label{subsec:benchmark}
Five state-of-the-art algorithms are compared in this section, namely: EHVI-MOBGO, PoI-MOBGO\footnote{Using EHVI or PoI as the infill criterion in Algorithm \ref{alg:mobgo}.}, NSGA-II \cite{Deb2002NSGA2}, NSGA-III \cite{Deb2014NSGA3,Platypus} and SMS-EMOA \cite{beume2007sms}. The benchmarks are DTLZ1, DTLZ2, DTLZ3, DTLZ4, DTLZ5, DTLZ7 \cite{dtlz2002a}, MaF1, MaF5, MaF12 and MaF13 \cite{Cheng2017}. The dimension $m$ of all the benchmarks is 6 or 18. The parameter settings for all of these algorithms are shown in Table \ref{tab:ParameterSettings}. The number of function evaluations ($T_c$ in Algorithm \ref{alg:mobgo}) for MOBGO based algorithms is 300. The Reference points for each benchmark are shown in Table \ref{tab:ReferencePointSettings}. For DTLZ1 and DTLZ2, the reference points are chosen from the article \cite{couckuyt2014fast}. For the other test problems, the reference points are revised from the articles \cite{couckuyt2014fast, Cheng2017}.
All experiments were repeated 10 times.
\begin{table}[!ht]
\centering
\caption{Algorithm Parameter Settings}
\label{tab:ParameterSettings}
\begin{tabular}{c|ccccc}
\bottomrule
                 & EHVI-MOBGO & PoI-MOBGO & NSGA-II  & NSGA-III & SMS-EMOA 		\\ 	\bottomrule
$\mu$            & 30       & 30      & 30       & /        & 30       		\\ 	
$\lambda$        & 1        & 1       & 30       &          & /        		\\ 	
Evaluation       & 300      & 300     & 300/2000 & 300/2000 & 300/2000 		\\ 	
divisions\_outer & /        & /       & /        & 12       & /        		\\	
$p_c$            & /        & /       & 0.9      &          & 0.9      		\\	
$p_m$            & /        & /       & 1/6      &          & 1/6      		\\	
Platform 		 & MATLAB 	& MATLAB  & MATLAB 	 & Python	& MATLAB		\\	\bottomrule
\end{tabular}
\end{table}

\begin{table}[!ht]
\centering
\caption{Reference Points}
\label{tab:ReferencePointSettings}
\begin{adjustbox}{width=1\textwidth}
\begin{tabular}{cccccccc}
\bottomrule
			 & DTLZ1      	 & DTLZ2 		 	 & DTLZ3 			 	& DTLZ4 			& DTLZ5 		& DTLZ7 	& MaF Problems\\ 	\bottomrule
$\mathbf{r}$ 
& (400,400,400) \cite{couckuyt2014fast} 
& (2.5,2.5,2.5)\cite{couckuyt2014fast}     
& (1500,1500,1500) 	
& (2.5,2.5,2.5) 	
& (11,11,11) 	
& (1,1,10)	
& (5,5,5)\\ 	\bottomrule
\end{tabular}
\end{adjustbox}
\end{table}

The final Pareto fronts are evaluated by means of the \emph{Hypervolume} indicator. 
%The experimental results (statistical mean and standard deviation) are shown in Table \ref{tab:EmpriricalComparisons} and Table \ref{tab:EmpriricalComparisons-18} for low-dimensional ($m=6$) and higher-dimensional ($m=18$) search spaces, respectively. 
Table \ref{tab:EmpriricalComparisons} and Table \ref{tab:EmpriricalComparisons-18} show the experimental results (statistical means and standard deviations) in 6- and 18-dimensional search spaces, respectively. 
%MOBGO based algorithms perform better than EMOAs (NSGA-II, NSGA-III and SMS-EMOA) with 300 function evaluations, sometimes even if the function evaluation budget of the EMOAs is increased to 2000.
Compared with EMOAs in 6- and 18-dimensional search spaces, either EHVI-MOBGO or PoI-MOBGO yields the best result on the 10 benchmark function with 300 function evaluations. 
On the test problems of DTLZ7 and MaF13, even if the function evaluation budget of the EMOAs is increased to 2000, EHVI-MOBGO can still outperform the EMOAs in 6- and 18-dimensional search spaces. 
%Among EHVI-MOBGO and PoI-MOBGO, EHVI-MOBGO outperforms PoI-MOBGO in most cases, except for DTLZ4 and DTLZ5. 
Among EHVI-MOBGO and PoI-MOBGO, EHVI-MOBGO outperforms PoI-MOBGO in most cases, but PoI-MOBGO yields better results on two and three (out of ten) test problems when $m=6$ and $m=18$, respectively. 
The reason is that the dominated space of the PoI is $\mathbf{y} \in (-\infty, \infty)^d \setminus \mbox{ dom }(\mathcal{P})$\footnote{A reference point in Equation (\ref{def:poi}) is defined as $\mathbf{r} = (-\infty, \infty, \infty)$ in this paper. For PoI-MOBGO, a reference point in Table \ref{tab:ReferencePointSettings} is only used to evaluate the experimental results.}, which is bigger than that of the EHVI $\mathbf{y} \in (\mathbf{r}, \infty^d) \setminus \mbox{ dom }(\mathcal{P})$. Therefore, PoI performs better when searching for extreme non-dominated points. In other words, EHVI is a reference based infill criterion and it cannot indicate any improvement of an evaluated point in a discarded part of the non-dominated space, namely, $\mathbf{y} \in (-\infty^d, \mathbf{r}) \setminus \mbox{ dom }(\mathcal{P})$.
%
%The reason is that $\mbox{PoI}$ is a reference-free indicator, and it considers all the possible improvement in a whole non-dominated space. However, EHVI only matters the above non-dominated space (in a maximization problem), which bounded by a reference point $\mathbf{r}$ and a Pareto-front approximation set $\pfa$. In other words, EHVI cannot indicate any improvement of an evaluated point in the non-dominated space, which is not bounded by a reference point. 

\begin{table}[!ht]
\centering
\caption{Empirical Comparisons w.r.t HV in the case of $m=6$}
\label{tab:EmpriricalComparisons}
\begin{adjustbox}{width=1\textwidth}
\begin{tabular}{ccc|ccccc||ccc}
\bottomrule
&	& 	& \multicolumn{2}{c}{MOBGO} 	& \multicolumn{3}{c||}{EAs} & \multicolumn{3}{c}{EAs} \\
\multicolumn{3}{c|}{Algorithm}             	& EHVI       & PoI    & NSGA-II       & NSGA-III      & SMS-EMOA    & NSGA-II       & NSGA-III      & SMS-EMOA   \\
\multicolumn{3}{c|}{Eval.}                 	& 300        & 300        & 300         & 300        & 300         & 2000        & 2000       & 2000       \\
% Problem	 	& Indicator 	& 	& 	& 	& 	& 	& 	& 	& 	& \\
\bottomrule
\multirow{2}{*}{DTLZ1} 
& \multirow{2}{*}{HV}  		& mean			& \bf{6.39587E+7} & 6.33975E+7 & 6.34583E+7  & 6.35749E+7 & 6.15372E+7  & 6.39914E+7  & \bf{6.39984E+7} & 6.39946E+7 \\
&							& std.					  & \bf{4.99505E+4} & 2.30970E+5 & 2.22275E+5  & 1.75929E+5 & 9.64758E+5  & 7.78434E+3  & \bf{2.01767E+3} & 5.68083E+3 \\
\bottomrule
\multirow{2}{*}{DTLZ2} 
& \multirow{2}{*}{HV}  		& mean         	& \bf{1.50203E+1} & 1.49975E+1 & 1.34232E+1  & 1.44291E+1 & 1.30529E+1  & 1.39295E+1  & \bf{1.49765E+1} & 1.46605E+1 \\
&							& std.                    & 1.02673E-2 & \bf{5.15545E-3} & 2.20199E-1  & 1.14701E-1 & 2.53151E-1  & 3.02290E-1  & \bf{5.18375E-3} & 9.54696E-2 \\
\bottomrule
\multirow{2}{*}{DTLZ3} 
& \multirow{2}{*}{HV}  		& mean          & \bf{3.37451E+9} & 3.36952E+9 & 3.35561E+9  & 3.36144E+9 & 3.28391E+9  & 3.37463E+9  & \bf{3.37499E+9} & 3.37459E+9 \\
&							& std.                    & \bf{1.41276E+5} & 4.75297E+6 & 8.29054E+6  & 3.54949E+6 & 2.51542E+7  & 2.75889E+5  & \bf{5.94791E+3} & 2.20550E+5 \\
\bottomrule
\multirow{2}{*}{DTLZ4}
& \multirow{2}{*}{HV}  		& mean          & 1.37964E+1 & \bf{1.44561E+1} & 1.24064E+1  & 1.22748E+1 & 9.64531E+0  & 1.31946E+1  & \bf{1.42249E+1} & 1.32491E+1 \\
&							& std.                    & 2.50980E-1 & \bf{1.60714E-1} & 1.36386E+0  & 5.77561E-1 & 7.48353E-1  & 1.60385E+0  & \bf{7.42880E-1} & 8.12735E-1 \\
\bottomrule
\multirow{2}{*}{DTLZ5} 
& \multirow{2}{*}{HV}  		& mean          & 1.31728E+3 & \bf{1.31883E+3} & 1.31089E+3  & 1.31549E+3 & 1.30092E+3  & 1.31742E+3  & \bf{1.31881E+3} & 1.31794E+3 \\
&							& std.                    & 3.69904E-1 & \bf{1.43867E-2} & 5.50786E+0  & 1.46033E+0 & 9.43520E+0  & 3.57031E-1  & \bf{4.52836E-2} & 3.43975E-1 \\
\bottomrule
\multirow{2}{*}{DTLZ7} 
& \multirow{2}{*}{HV}  		& mean          & \bf{5.08646E+0} & 4.06894E+0 & 2.45578E+0  & 1.59954E-1 & 2.43627E+0 & \bf{4.96201E+0} & 4.91579E+0 & 1.06682E-1 \\
&							& std.                    & \bf{7.59397E-2} & 1.42676E-1 & 7.00249E-1  & 5.81027E-1 & 7.96762E-1 & \bf{1.14735E-1} & 1.61886E-1 & 7.87267E-1 \\
\bottomrule
\multirow{2}{*}{MaF1} 
& \multirow{2}{*}{HV}  		& mean    		& \bf{1.17445E+2} & 1.17153E+2 & 1.14050E+2  & 1.05551E+2 & 1.11801E+2 & \bf{1.16459E+2} & {1.16079E+2} & 1.13299E+2 \\
&						 	& std.                    & \bf{4.96840E-2} & 6.87363E-2 & 1.05600E+0  & 1.45637E+0 & 1.54941E+0 & \bf{1.42786E-1} & {4.47650E-1} & 1.19772E+0 \\
\bottomrule
%\multirow{2}{*}{MaF3} 
%& mean                    & \bf{5.08646E+0} & 4.06894E+0 & 0  & 1.59954E-1 & 0 & 0 & \bf{4.91579E+0} & 1.06682E-1 \\
%& std.                    & \bf{7.59397E-2} & 1.42676E-1 & /  & 5.81027E-1 & / &  / & \bf{1.61886E-1} & 7.87267E-1 \\
%\bottomrule
\multirow{2}{*}{MaF5} 
& \multirow{2}{*}{HV}  		& mean         	& \bf{7.23424E+1} & 6.76919E+1 & 5.62684E+1  & 4.27219E+1 & 2.75989E+1 & \bf{8.61379E+1} & {8.13403E+1} & 8.32843E+1 \\
&							& std.                    & \bf{1.32736E+1} & 2.76597E+1 & 2.07352E+1  & 1.83150E+1 & 3.10755E+1 & \bf{1.98541E+1} & {2.01754E+1} & 1.99339E+1 \\
\bottomrule
\multirow{2}{*}{MaF12} 
& \multirow{2}{*}{HV}  		& mean   		& \bf{8.35314E+1} & 8.20753E+1 & 7.05794E+1  & 5.97979E+1 & 7.07943E+1 & 9.17876E+1 & {8.33820E+1} & \bf{9.40144E+1} \\
&							& std.                    & \bf{3.47560E+0} & 8.10423E+0 & 4.83604E+0  & 3.53278E+0 & 4.09591E+0 & 8.85050E-1 & {3.21738E+0} & \bf{4.42730E-1} \\
\bottomrule
\multirow{2}{*}{MaF13} 
& \multirow{2}{*}{HV}  		& mean   		& \bf{1.24153E+2} & 1.22594E+2 & 1.07438E+2  & 1.13489E+2 & 1.09264E+2 & 1.22595E+2 & \bf{1.23203E+2} & 1.22414E+2 \\
&							& std.                    & \bf{3.02702E-1} & 7.76629E-1 & 5.69950E+0  & 3.83089E+0 & 4.07713E+0 & 7.81655E-1 & \bf{5.29024E-1} & 9.04852E-1 \\
\bottomrule
\end{tabular}
\end{adjustbox}
\end{table}

\begin{table}[!ht]
\centering
\caption{Empirical Comparisons w.r.t HV in the case of $m=18$}
\label{tab:EmpriricalComparisons-18}
\begin{adjustbox}{width=1\textwidth}
\begin{tabular}{ccc|ccccc||ccc}
\bottomrule
&	& 	& \multicolumn{2}{c}{MOBGO} 	& \multicolumn{3}{c||}{EAs} & \multicolumn{3}{c}{EAs} \\
\multicolumn{3}{c|}{Algorithm}             	& EHVI       & PoI    & NSGA-II       & NSGA-III      & SMS-EMOA    & NSGA-II       & NSGA-III      & SMS-EMOA   \\
\multicolumn{3}{c|}{Eval.}                 	& 300        & 300        & 300         & 300        & 300         & 2000        & 2000       & 2000       \\
% Problem	 	& Indicator 	& 	& 	& 	& 	& 	& 	& 	& 	& \\
\bottomrule
\multirow{2}{*}{DTLZ1} 
& \multirow{2}{*}{HV}  		& mean			& \bf{5.1288E+7}  & 2.7077E+7 & 2.2253E+7  & 2.3745E+7 & 1.8500E+7 & 5.9953E+7 & \bf{6.3317E+7} & 5.9134E+7 \\
&							& std.			& \bf{3.0564E+6}  & 7.8473E+6 & 4.6439E+6  & 3.3495E+6 & 6.9285E+6 & 1.0099E+6 & \bf{2.5485E+5} & 1.5575E+6 \\
\bottomrule
\multirow{2}{*}{DTLZ2} 
& \multirow{2}{*}{HV}  		& mean         	& 1.2639E+1  & \bf{1.4239E+1} & 8.5169E+0 & 1.2543E+1 & 9.2854E+0 & 1.0758E+1 & \bf{1.4859E+1} & 1.2128E+1 \\
&							& std.          & 1.8477E+0  & 3.9014E-1 & 1.0017E+0 & \bf{3.8405E-1} & 6.2299E-1 & 5.8787E-1 & \bf{3.6867E-2} & 4.2366E-1 \\
\bottomrule
\multirow{2}{*}{DTLZ3} 
& \multirow{2}{*}{HV}  		& mean          & \bf{3.2911E+9} & 2.6798E+9  & 2.1654E+9 & 2.6382E+9 & 2.2338E+9 & 3.2974E+9 & \bf{3.3655E+9} & 3.2967E+9 \\
&							& std.          & \bf{1.1090E+7} & 1.8750E+8  & 1.9180E+8 & 1.7309E+8 & 2.6704E+8 & 2.4809E+8 & \bf{3.8824E+6} & 3.2490E+8 \\
\bottomrule
\multirow{2}{*}{DTLZ4}
& \multirow{2}{*}{HV}  		& mean          & 8.3732E+0 & \bf{1.2054E+1} & 6.5989E+0 & 7.5245E+0 & 7.1119E+0 & 7.9902E+0 & \bf{1.3305E+1} & 9.2049E+0 \\
&							& std.          & 2.7638E+0 & \bf{1.1977E+0} & 1.3956E+0 & 1.4967E+0 & 1.9694E+0 & \bf{1.7487E+0} & 2.2255E+0
 & 2.0953E+0 \\
\bottomrule
\multirow{2}{*}{DTLZ5} 
& \multirow{2}{*}{HV}  		& mean          & \bf{1.3128E+3} & 1.3050E+3  & 1.2852E+3 & 1.3027E+3 & 1.2883E+3 & 1.3042E+3 & \bf{1.3176E+3} & 1.3077E+3  \\
&							& std.          & \bf{1.5919E+0} & 1.7630E+0  & 1.0882E+1 & 4.0443E+0 & 1.4893E+1 & 2.6870E+0 & \bf{8.5620E-1} & 1.9797E+0  \\
\bottomrule
\multirow{2}{*}{DTLZ7} 
& \multirow{2}{*}{HV}  		& mean          & \bf{4.5911E+0} & 1.0198E+0 & 0 & 0 & 0 & 0 & \bf{3.5493E+0} & 3.8917E-3 \\
&							& std.          & \bf{6.1310E-1} & 8.4290E-1 & / & / & / & / & \bf{2.4150E-1} & 1.2307E+2 \\
\bottomrule
\multirow{2}{*}{MaF1} 
& \multirow{2}{*}{HV}  		& mean    		& \bf{1.0473E+2}  & 1.0381E+2 & 6.7074E+1 & 9.9441E+1 & 8.8914E+1 & 9.6273E+1 & \bf{1.1563E+2} & 1.0034E+2 \\
&						 	& std.          & 4.4597E+0 & \bf{2.4946E+0} & 3.5408E+0 & 2.0910E+0 & 2.7782E+0 & 3.5475E+0 & \bf{3.6416E-1} & 2.2847E+0 \\
\bottomrule
%\multirow{2}{*}{MaF3} 
%& mean                    & \bf{5.08646E+0} & 4.06894E+0 & 0  & 1.59954E-1 & 0 & 0 & \bf{4.91579E+0} & 1.06682E-1 \\
%& std.                    & \bf{7.59397E-2} & 1.42676E-1 & /  & 5.81027E-1 & / &  / & \bf{1.61886E-1} & 7.87267E-1 \\
%\bottomrule
\multirow{2}{*}{MaF5} 
& \multirow{2}{*}{HV}  		& mean         	& \bf{3.8302E+1} & 2.9227E+1 & 6.7114E+0 & 2.4487E+1 & 1.0390E+1 & 2.0824E+1 & \bf{8.1290E+1} & 1.6117E+1 \\
&							& std.          & \bf{1.4669E+1} & 2.1667E+1 & 1.3320E+1 & 2.2095E+1 & 1.4802E+1 & 2.0555E+1 & 2.0022E+1 & \bf{1.9612E+1} \\
\bottomrule
\multirow{2}{*}{MaF12} 
& \multirow{2}{*}{HV}  		& mean   		& \bf{7.1659E+1} & 6.3981E+1 	& 4.2264E+1  & 5.6579E+1 & 5.1103E+1 & 5.4358E+1 & \bf{7.9808E+1} & 6.5401E+1\\
&							& std.          & 1.2235E+1 & 1.9204E+1  & 7.8510E+0  & \bf{3.4781E+0} & 6.6979E+0 & 6.5554E+0 & \bf{3.2876E+0} & 3.8870E+0\\
\bottomrule
\multirow{2}{*}{MaF13} 
& \multirow{2}{*}{HV}  		& mean   		& 9.5580E+1 & \bf{1.2260E+2} 	& 3.1497E+1  & 6.9621E+1 & 3.9234E+1 & 5.4296E+1 & \bf{1.1731E+2} & 6.8469E+1 \\
&							& std.          & 2.7625E+1 & \bf{1.1757E+0} 	& 6.8408E+0  & 6.2208E+0 & 6.4776E+0 & 9.4731E+0 & \bf{1.9190E+0} & 7.4302E+0\\
\bottomrule
\end{tabular}
\end{adjustbox}
\end{table}
	
\section{Conclusions and Outlook}\label{sec:Conclusions} 
This paper describes an efficient algorithm for EHVI calculation. It reviews and benchmarks recently proposed asymptotically optimal algorithms with $\Theta(n\log n)$ time complexity and generalizes them to higher dimensional cases with $d\geq4$.
%The new integration technique is able to make use of the fast box decomposition techniques for the non-dominated space that were recently developed by D{\"a}chert et al. \cite{dachert2017efficient} and Lacour et al. \cite{lacour2017box}, that partition the non-dominated space into only $\Theta(n^{\lfloor d/2 \rfloor})$ hyperboxes. 
By using the fast box decomposition techniques, which were recently developed by D{\"a}chert et al. \cite{dachert2017efficient} and Lacour et al. \cite{lacour2017box}, a non-dominated space can be partitioned with only $\Theta(n^{\lfloor d/2 \rfloor})$ hyperboxes. 
The time complexity of our new EHVI computation algorithm scales linearly with the number of hyperboxes of arbitrary box decompositions. Unlike previous EHVI computation algorithms, the new algorithm does not require full grid partitionings with $\Theta(n^d)$ boxes. The new algorithm is, therefore, a significant improvement in terms of asymptotic time complexity. In addition, our benchmarks on random non-dominated sets show that this new algorithm is also many orders of magnitude faster for computations of typically sized problems where $n \leq 1000$. 

%So far, the speed of fast EHVI implementations has only been tested for $d\leq 3$ and on random non-dominated sets.
%This paper also compares the speed and performance of fast EHVI-MOBGO with four other state-of-the-art multi-objective optimization algorithms on benchmark problems in multi-objective optimization. 
This paper also compares the performance of MOBGO based algorithms with three other state-of-the-art  EMOAs on 10 benchmark test problems. 
For budgets of function evaluations up to 300, MOBGO based algorithms can yield the Pareto-front approximation sets with higher HV values than that of EMOAs in both 6- and 18-dimensional search spaces. 
%Compared with PoI-MOBGO, the Pareto-front approximation sets generated by EHVI-MOBGO are usually closer to the true Pareto front on the tested 10 benchmark functions. 
%However, PoI-MOBGO performs better than EHVI-MOBGO on DTLZ4 and DTLZ5 problems. 
In most cases, EHVI-MOBGO yields better performance than PoI-MOBGO. However, PoI-MOBGO performs better than EHVI-MOBGO on up to 3 test problems in 6- and 18- dimensional search spaces. 
%The reason is that PoI is a reference-free criterion and EHVI is a reference-based criterion. 
%PoI implies an improvement of an evaluated point in the whole non-dominated space, but the EHVI can only indicate the improvement in the non-dominated space bounded by a reference point. 
The reason is that the PoI can imply an improvement of an evaluated point in the whole non-dominated space, but the EHVI can only indicate the improvement in the non-dominated space bounded by a reference point. 
A remedy to the EHVI's disadvantage can be achieved by setting a large reference point or using the dynamic reference point strategy. 
However, the selected reference point must not be too large. 
Otherwise, EHVI at any evaluated points would be similar, even identical, because of the insufficient numerical stability of the computations involved. 

For future research, it is recommended to further investigate on reference-free computation of EHVI. Moreover, it is still an open question of how to obtain fast EHVI calculations for a larger number of objective functions. Although it is conjectured that in the worst case time complexity will increase superpolynomially with the number of objectives, a better average case time complexity could be obtained by adopting concepts from recently proposed divide-and-conquer algorithms for computing the hypervolume indicator \cite{russo2014quick,JASZKIEWICZ201872}. 

\begin{acknowledgements}
The authors gratefully acknowledge the contribution of Carlos M. Fonseca (University of Coimbra, Portugal) on the discussion of the integration techniques.
\end{acknowledgements}
%%
%%\bibliographystyle{spmpsci}      % mathematics and physical sciences
%%%\bibliographystyle{spphys}       % APS-like style for physics
%%%\bibliography{}   % name your BibTeX data base
%%%\bibliography{kaifeng}
%%\bibliography{Kaifeng-references}

\begin{thebibliography}{10}
\providecommand{\url}[1]{{#1}}
\providecommand{\urlprefix}{URL }
\expandafter\ifx\csname urlstyle\endcsname\relax
  \providecommand{\doi}[1]{DOI~\discretionary{}{}{}#1}\else
  \providecommand{\doi}{DOI~\discretionary{}{}{}\begingroup
  \urlstyle{rm}\Url}\fi

\bibitem{beume2007sms}
Beume, N., Naujoks, B., Emmerich, M.: {S}{M}{S}-{E}{M}{O}{A}: Multiobjective
  selection based on dominated hypervolume.
\newblock European Journal of Operational Research \textbf{181}(3), 1653--1669
  (2007)

\bibitem{Cheng2017}
Cheng, R., Li, M., Tian, Y., Zhang, X., Yang, S., Jin, Y., Yao, X.: A benchmark
  test suite for evolutionary many-objective optimization.
\newblock Complex {\&} Intelligent Systems \textbf{3}(1), 67--81 (2017).
\newblock \doi{10.1007/s40747-017-0039-7}.
\newblock \urlprefix\url{https://doi.org/10.1007/s40747-017-0039-7}

\bibitem{chugh2017handling}
Chugh, T.: Handling expensive multiobjective optimization problems with
  evolutionary algorithms.
\newblock Ph.D. thesis, Faculty of Information Technology, University of
  Jyv{\"a}skyl{\"a} (2017)

\bibitem{CoelloCoello2011}
Coello~Coello, C.A.: Evolutionary multi-objective optimization: Basic concepts
  and some applications in pattern recognition.
\newblock In: J.F. Mart{\'i}nez-Trinidad, J.A. Carrasco-Ochoa,
  C.~Ben-Youssef~Brants, E.R. Hancock (eds.) Proceedings of the Third Mexican
  conference on Pattern recognition, pp. 22--33. Springer, Berlin, Heidelberg
  (2011).
\newblock \doi{10.1007/978-3-642-21587-2_3}.
\newblock \urlprefix\url{http://dx.doi.org/10.1007/978-3-642-21587-2_3}

\bibitem{couckuyt2014fast}
Couckuyt, I., Deschrijver, D., Dhaene, T.: Fast calculation of multiobjective
  probability of improvement and expected improvement criteria for {P}areto
  optimization.
\newblock Journal of Global Optimization \textbf{60}(3), 575--594 (2014)

\bibitem{dachert2017efficient}
D{\"a}chert, K., Klamroth, K., Lacour, R., Vanderpooten, D.: Efficient
  computation of the search region in multi-objective optimization.
\newblock European Journal of Operational Research \textbf{260}(3), 841--855
  (2017)

\bibitem{Deb2014NSGA3}
Deb, K., Jain, H.: An evolutionary many-objective optimization algorithm using
  reference-point-based nondominated sorting approach, {P}art {I}: Solving
  problems with box constraints.
\newblock IEEE Transactions on Evolutionary Computation \textbf{18}(4),
  577--601 (2014).
\newblock \doi{10.1109/TEVC.2013.2281535}

\bibitem{Deb2002NSGA2}
Deb, K., Pratap, A., Agarwal, S., Meyarivan, T.: A fast and elitist
  multiobjective genetic algorithm: {NSGA-II}.
\newblock IEEE Transactions on Evolutionary Computation \textbf{6}(2), 182--197
  (2002).
\newblock \doi{10.1109/4235.996017}

\bibitem{dtlz2002a}
Deb, K., Thiele, L., Laumanns, M., Zitzler, E.: Scalable multi-objective
  optimization test problems.
\newblock In: Proceedings of the 2002 Congress on Evolutionary Computation.
  CEC'02 (Cat. No.02TH8600), vol.~1, pp. 825--830 vol.1 (2002).
\newblock \doi{10.1109/CEC.2002.1007032}

\bibitem{emmerich2005single}
Emmerich, M.: Single-and multi-objective evolutionary design optimization
  assisted by {G}aussian random field metamodels.
\newblock Ph.D. thesis, Fachbereich Informatik, Chair of Systems Analysis,
  University of Dortmund (2005)

\bibitem{emmerich2008computation}
Emmerich, M., Deutz, A., Klinkenberg, J.W.: The computation of the expected
  improvement in dominated hypervolume of {P}areto front approximations.
\newblock Technical Report, Leiden University \textbf{34} (2008)

\bibitem{emmerich2011hypervolume}
Emmerich, M., Deutz, A.H., Klinkenberg, J.W.: Hypervolume-based expected
  improvement: Monotonicity properties and exact computation.
\newblock In: 2011 IEEE Congress on Evolutionary Computation (CEC), pp.
  2147--2154. IEEE (2011)

\bibitem{emmerich2006single}
Emmerich, M., Giannakoglou, K.C., Naujoks, B.: Single-and multiobjective
  evolutionary optimization assisted by {G}aussian random field metamodels.
\newblock IEEE Transactions on Evolutionary Computation \textbf{10}(4),
  421--439 (2006)

\bibitem{Michael2016book}
Emmerich, M., Yang, K., Deutz, A., Wang, H., Fonseca, C.M.: A multicriteria
  generalization of Bayesian global optimization.
\newblock In: P.M. Pardalos, A.~Zhigljavsky, J.~{\v{Z}}ilinskas (eds.) Advances
  in Stochastic and Deterministic Global Optimization, pp. 229--243. Springer,
  Berlin, Heidelberg (2016)

\bibitem{emmerich2011computing}
Emmerich, M.T.M., Fonseca, C.M.: Computing hypervolume contributions in low
  dimensions: Asymptotically optimal algorithm and complexity results.
\newblock In: R.H.C. Takahashi, K.~Deb, E.F. Wanner, S.~Greco (eds.)
  International Conference on Evolutionary Multi-Criterion Optimization, pp.
  121--135. Springer, Berlin, Heidelberg (2011)

\bibitem{gaida2014dynamic}
Gaida, D.: Dynamic real-time substrate feed optimization of anaerobic
  co-digestion plants.
\newblock Ph.D. thesis, Leiden Institute of Advanced Computer Science (LIACS),
  Faculty of Science, Leiden University (2014)

\bibitem{Platypus}
Hadka, D.: Platypus - {M}ultiobjective optimization in {P}ython.
\newblock pp. https://github.com/Project--Platypus/Platypus (2015)

\bibitem{hansen2009benchmarking}
Hansen, N.: Benchmarking a {BI}-population {CMA-ES} on the {BBOB}-2009 function
  testbed.
\newblock In: Proceedings of the 11th Annual Conference Companion on Genetic
  and Evolutionary Computation Conference: Late Breaking Papers (GECOO), pp.
  2389--2396. ACM, New York, USA (2009).
\newblock \doi{10.1145/1570256.1570333}.
\newblock \urlprefix\url{http://doi.acm.org/10.1145/1570256.1570333}

\bibitem{hupkens2015faster}
Hupkens, I., Deutz, A., Yang, K., Emmerich, M.: Faster exact algorithms for
  computing expected hypervolume improvement.
\newblock In: A.~Gaspar-Cunha, C.~Henggeler~Antunes, C.C. Coello (eds.)
  International Conference on Evolutionary Multi-Criterion Optimization, pp.
  65--79. Springer, Cham (2015)

\bibitem{JASZKIEWICZ201872}
Jaszkiewicz, A.: Improved quick hypervolume algorithm.
\newblock Computers \& Operations Research \textbf{90}, 72 -- 83 (2018).
\newblock \doi{https://doi.org/10.1016/j.cor.2017.09.016}.
\newblock
  \urlprefix\url{http://www.sciencedirect.com/science/article/pii/S0305054817302484}

\bibitem{jones1998efficient}
Jones, D.R., Schonlau, M., Welch, W.J.: Efficient global optimization of
  expensive black-box functions.
\newblock Journal of Global Optimization \textbf{13}(4), 455--492 (1998)

\bibitem{keane2006statistical}
Keane, A.J.: Statistical improvement criteria for use in multiobjective design
  optimization.
\newblock American Institute of Aeronautics and Astronautics (AIAA) Journal
  \textbf{44}(4), 879--891 (2006)

\bibitem{koch2015efficientnoise}
Koch, P., Wagner, T., Emmerich, M.T., B{\"a}ck, T., Konen, W.: Efficient
  multi-criteria optimization on noisy machine learning problems.
\newblock Applied Soft Computing \textbf{29}, 357 -- 370 (2015).
\newblock \doi{https://doi.org/10.1016/j.asoc.2015.01.005}.
\newblock
  \urlprefix\url{http://www.sciencedirect.com/science/article/pii/S156849461500006X}

\bibitem{kushner1964poi}
Kushner, H.J.: A new method of locating the maximum point of an arbitrary
  multi-peak curve in the presence of noise.
\newblock Journal of Basic Engineering \textbf{86}(1), 97--106 (1964)

\bibitem{lacour2017box}
Lacour, R., Klamroth, K., Fonseca, C.M.: A box decomposition algorithm to
  compute the hypervolume indicator.
\newblock Computers \& Operations Research \textbf{79}, 347--360 (2017)

\bibitem{lagarias1998fminsearch}
Lagarias, J., Reeds, J., Wright, M., Wright, P.: Convergence properties of the
  nelder--mead simplex method in low dimensions.
\newblock SIAM Journal on Optimization \textbf{9}(1), 112--147 (1998).
\newblock \doi{10.1137/S1052623496303470}.
\newblock \urlprefix\url{https://doi.org/10.1137/S1052623496303470}

\bibitem{laniewski2010development}
{\L}aniewski-Wo{\l}{\l}k, {\L}., Obayashi, S., Jeong, S.: Development of
  expected improvement for multi-objective problems.
\newblock In: Proceedings of 42nd Fluid Dynamics Conference/Aerospace
  Numerical, Simulation Symposium (CD ROM). Varna, Bulgaria (2010)

\bibitem{li2008metamodel}
Li, R., Emmerich, M.T.M., Eggermont, J., Bovenkamp, E.G.P., B{\"a}ck, T.,
  Dijkstra, J., Reiber, J.H.C.: Metamodel-assisted mixed integer evolution
  strategies and their application to intravascular ultrasound image analysis.
\newblock In: 2008 IEEE Congress on Evolutionary Computation (IEEE World
  Congress on Computational Intelligence), pp. 2764--2771 (2008).
\newblock \doi{10.1109/CEC.2008.4631169}

\bibitem{lhs1979}
McKay, M.D., Beckman, R.J., Conover, W.J.: Comparison of three methods for
  selecting values of input variables in the analysis of output from a computer
  code.
\newblock Technometrics \textbf{21}(2), 239--245 (1979).
\newblock \doi{10.1080/00401706.1979.10489755}.
\newblock \urlprefix\url{https://doi.org/10.1080/00401706.1979.10489755}

\bibitem{Mockus1978}
Mockus, J., Tie{\v{s}}is, V., {\v{Z}}ilinskas, A.: {The application of Bayesian
  methods for seeking the extremum}.
\newblock In: L.~Dixon, G.~Szeg{\"{o}} (eds.) Towards Global Optimization,
  vol.~2, pp. 117--131. North-Holland, Amsterdam (1978)

\bibitem{russo2014quick}
Russo, L.M.S., Francisco, A.P.: Quick hypervolume.
\newblock IEEE Transactions on Evolutionary Computation \textbf{18}(4),
  481--502 (2014).
\newblock \doi{10.1109/TEVC.2013.2281525}

\bibitem{shimoyama2013kriging}
Shimoyama, K., Jeong, S., Obayashi, S.: Kriging-surrogate-based optimization
  considering expected hypervolume improvement in non-constrained
  many-objective test problems.
\newblock In: 2013 IEEE Congress on Evolutionary Computation, pp. 658--665
  (2013).
\newblock \doi{10.1109/CEC.2013.6557631}

\bibitem{shimoyama2013updating}
Shimoyama, K., Sato, K., Jeong, S., Obayashi, S.: Updating Kriging surrogate
  models based on the hypervolume indicator in multi-objective optimization.
\newblock Journal of Mechanical Design \textbf{135}(9), 094,503 (2013)

\bibitem{stuckman1988global}
Stuckman, B.E.: A global search method for optimizing nonlinear systems.
\newblock IEEE Transactions on Systems, Man, and Cybernetics \textbf{18}(6),
  965--977 (1988)

\bibitem{ulmer2003evolution}
Ulmer, H., Streichert, F., Zell, A.: Evolution strategies assisted by Gaussian
  processes with improved preselection criterion.
\newblock In: The 2003 Congress on Evolutionary Computation, 2003. CEC '03.,
  vol.~1, pp. 692--699 Vol.1 (2003).
\newblock \doi{10.1109/CEC.2003.1299643}

\bibitem{wagner2010expected}
Wagner, T., Emmerich, M., Deutz, A., Ponweiser, W.: On expected-improvement
  criteria for model-based multi-objective optimization.
\newblock In: R.~Schaefer, C.~Cotta, J.~Ko\l{}odziej, G.~Rudolph (eds.)
  International Conference on Parallel Problem Solving from Nature--PPSN XI,
  pp. 718--727. Springer, Berlin, Heidelberg (2010)

\bibitem{kyang2016cec}
Yang, K., Deutz, A., Yang, Z., B{\"a}ck, T., Emmerich, M.: Truncated expected
  hypervolume improvement: Exact computation and application.
\newblock In: 2016 IEEE Congress on Evolutionary Computation (CEC), pp.
  4350--4357. IEEE (2016).
\newblock \doi{10.1109/CEC.2016.7744343}

\bibitem{yang2017computing}
Yang, K., Emmerich, M., Deutz, A., Fonseca, C.M.: Computing 3-{D} expected
  hypervolume improvement and related integrals in asymptotically optimal time.
\newblock In: H.~Trautmann, G.~Rudolph, K.~Klamroth, O.~Sch\"{u}tze, Y.~Wiecek
  Margaret~Jin, C.~Grimme (eds.) International Conference on Evolutionary
  Multi-Criterion Optimization, pp. 685--700. Springer, Cham (2017)

\bibitem{yang2017phdthesis}  
Yang, K.: Multi-Objective Bayesian Global Optimization for Continuous Problems and Applications. 
\newblock Ph.D. thesis, Leiden Institute of Advanced Computer Science (LIACS), Faculty of Science, Leiden University (2017)

\bibitem{kyang2015cec}
Yang, K., Gaida, D., B{\"a}ck, T., Emmerich, M.: Expected hypervolume
  improvement algorithm for {PID} controller tuning and the multiobjective
  dynamical control of a biogas plant.
\newblock In: 2015 IEEE Congress on Evolutionary Computation (CEC), pp.
  1934--1942. IEEE (2015).
\newblock \doi{10.1109/CEC.2015.7257122}

\bibitem{zaefferer2013case}
Zaefferer, M., Bartz-Beielstein, T., Naujoks, B., Wagner, T., Emmerich, M.: A
  case study on multi-criteria optimization of an event detection software
  under limited budgets.
\newblock In: R.C. Purshouse, P.J. Fleming, C.M. Fonseca, S.~Greco, J.~Shaw
  (eds.) International Conference on Evolutionary Multi-Criterion Optimization,
  pp. 756--770. Springer, Berlin, Heidelberg (2013)

\bibitem{zitzler1999multiobjective}
Zitzler, E., Thiele, L.: Multiobjective evolutionary algorithms: a comparative
  case study and the strength {P}areto approach.
\newblock IEEE Transactions on Evolutionary Computation \textbf{3}(4), 257--271
  (1999)

\bibitem{zitzler2003performance}
Zitzler, E., Thiele, L., Laumanns, M., Fonseca, C.M., Da~Fonseca, V.G.:
  Performance assessment of multiobjective optimizers: an analysis and review.
\newblock Evolutionary Computation, IEEE Transactions on \textbf{7}(2),
  117--132 (2003)
 


\end{thebibliography}

%\begin{thebibliography}{}
%%
%% and use \bibitem to create references. Consult the Instructions
%% for authors for reference list style.
%\end{thebibliography}

\end{document}